\documentclass[runningheads]{llncs}
\usepackage{graphicx}
\usepackage{comment}
\usepackage{amsmath,amssymb} 
\usepackage{color}

\usepackage{helvet} 
\usepackage{courier}  
\usepackage[hyphens]{url}  
\usepackage{subcaption} 
\usepackage{xparse}
\usepackage[font=small,labelfont=bf]{caption}

\usepackage[draft]{hyperref}       
\usepackage{url}            
\usepackage{amsfonts}       
\usepackage{nicefrac}       
\usepackage{microtype}      

\usepackage{epstopdf}
\usepackage{rotating,booktabs,multirow}
\usepackage{wrapfig,lipsum}
\usepackage{tabularx}

\usepackage{algorithm}
\usepackage{algorithmic}
\usepackage[algo2e]{algorithm2e}
\usepackage{mathtools}

\usepackage[width=122mm,left=12mm,paperwidth=146mm,height=193mm,top=12mm,paperheight=217mm]{geometry}

\newcommand{\etal}{\textit{et al}. }
\newcommand{\ie}{\textit{i}.\textit{e}., }
\newcommand{\eg}{\textit{e}.\textit{g}. }

\begin{document}
\pagestyle{headings}
\mainmatter
\def\ECCVSubNumber{4648}  

\title{Predictively Encoded Graph Convolutional Network for Noise-Robust Skeleton-based Action Recognition} 

\titlerunning{ECCV-20 submission ID \ECCVSubNumber} 
\authorrunning{ECCV-20 submission ID \ECCVSubNumber} 
\author{Jongmin Yu$^{1,2}$, Yongsang Yoon$^{2}$, and Moongu Jeon$^{2}$}
\institute{Curtin University, WA, Australia\\Gwangju Institution of Science and Technology, Gwangju, South Korea\\
jm.andrew.yu@gmail.com$^{1,2}$, \{nil, mgjeon\}@gist.ac.kr}

\maketitle

\begin{abstract}
In skeleton-based action recognition, graph convolutional networks (GCNs), which model human body
skeletons using graphical components such as nodes and connections, have achieved remarkable performance recently. However, current state-of-the-art methods for skeleton-based action recognition usually work on the assumption that the completely observed skeletons will be provided. This may be problematic to apply this assumption in real scenarios since there is always a possibility that captured skeletons are incomplete or noisy. In this work, we propose a skeleton-based action recognition method which is robust to noise information of given skeleton features. The key insight of our approach is to train a model by maximizing the mutual information between normal and noisy skeletons using a predictive coding manner. We have conducted comprehensive experiments about skeleton-based action recognition with defected skeletons using NTU-RGB+D and Kinetics-Skeleton datasets. The experimental results demonstrate that our approach achieves outstanding performance when skeleton samples are noised compared with existing state-of-the-art methods.
\end{abstract}

\section{Introduction}
Action recognition is one of the important areas in computer vision studies, for understanding human behaviours using a computational system. It can be applied to various applications for industrial system \cite{tran20113,nozaki2012recognition}, medical software \cite{bahrepour2011sensor}, and multimedia \cite{zhang2012microsoft,wang2017two}. Because of the industrial and practical importance of this literature, the interest of this literature is increasing rapidly in recent years, and numerous studies have been proposed. In general, various modalities, such as appearance \cite{Feichtenhofer_2016_CVPR}, depth \cite{zhang2016rgb,liu2019learning}, motion flow \cite{wang2019hallucinating}, and skeleton-features \cite{si2019attention} are utilized to recognize human actions. With the great advancements of deep learning which is a method to learn useful representation automatically, various approaches employ convolutional neural networks (CNNs) \cite{kim2017interpretable,ding2017investigation,carreira2017quo} and recurrent neural networks (RNNs) \cite{liu2016spatio,zhang2018fusing,zhang2017view} to train the spatio-temporal information and to recognize human actions. These CNNs and RNNs based approaches using RGB images and motion flows (\eg, optical flow) achieved outstanding performances than the previous methods based on hand-crafted features \cite{wang2013action,xia2012view}. The drawback of these approaches is that the learnt representations are may not focused on actions since entire areas of video frames are exploited to learn the representations \cite{shi2019two,fernando2015modeling}. Skeleton features provide quantized information about peoples' joints and bones. Compared to RGBs and motion flows, the skeleton features can provide more compact and useful information in the dynamic circumstance and complicated background \cite{vemulapalli2014human,fernando2015modeling,du2015hierarchical,ke2017new}. 

Early deep-learning based approaches using skeleton-features manually create skeleton data as a sequence of joint-coordinate vectores \cite{du2015hierarchical,shahroudy2016ntu,liu2016spatio,song2017end,zhang2017view} or as a pesudo-image \cite{ke2017new,kim2017interpretable,liu2017enhanced}, and apply the data to RNNs or CNNs to inference corresponding action classes. However, these approaches are unable to indicate the dependency between correlated joints \cite{shi2019two}. Intuitively, skeleton-features can be represented as a graph structure since their components are homeomorphic. For instance, joints and bones of skeleton-features can be defined as the vertices and connections of the graph. Recently, Graph Convolutional Networks (GCNs), which are the graphical framework using convolution neural network, have been achieved a great number of successes in skeleton-based action recognition \cite{shi2019two,si2019attention,li2019actional}. In the graph convolution, nodes are filtered in two ways, namely spatial and spectral. Spectral approach filters the nodes based on the Laplacian matrix and eigenvectors while spatial approach filters the nodes with local neighborhood nodes. ST-GCN \cite{yan2018spatial} is the first work to use spatial approach GCNs to handle skeleton model and has shown impressive improvements. However, the spatial graph in ST-GCN is a predefined graph which only relies on the physical structure of human body. This makes hard to capture the relationship between closely related joints such as both two hands in hand-related action. To tackle this limitation, many methods \cite{shi2019two,shi2019skeleton,song2019richly,li2019actional,si2019attention} were proposed to build adaptive graph to pay the dynamic attention to each joint based on the performing action.

Unfortunately, all these approaches assume that the complete skeleton features would be provided. It is almost impossible to guarantee that extracting of perfect skeleton samples from a real-world system. Song \etal \cite{song2019richly} have proposed a GCN based method which can deal with '\textit{incomplete skeletons}' defined as spatially occluded or temporally missed skeleton features. Even though recent studies for pose estimation \cite{cao2018openpose,wang2019densefusion} and constructing skeleton-features \cite{ding2017investigation,ke2017new,liu2017enhanced}, have shown precise and scene-condition invariant performances, still, there is a possibility that extraed skeleton may contain a piece of inaccurate information. Song \etal proposed RA-GCN \cite{song2019richly} model to learn distinctive features of currently unactivated joints (missed joints) in multiple streams by utilizing class activation maps (CAM), but it is still problematic. To improve the performance of action recognition using skeleton-features, it should have addressed how a model processes noisy skeleton samples. 

We present Predictively encoded Graph Convolutional Networks (PeGCNs), which can learn noise-robust representation for skeleton-based action recognition using GCN. The key insight of our model is to learn such representations by predicting the perfect sample from noisy sample in latent space via autoregression model. We use a probabilistic contrastive loss to capture the most useful information for predicting a perfect sample. To demonstrate the efficiency of PeGCNs on skeleton-based action recogntion with noised samples, we have conducted various experiments using NTU-RGB+D \cite{shahroudy2016ntu} and Kinetics-skeleton \cite{yan2018spatial} datasets. The experimental results show that PeGCNs can provide noise-robust action recognition performance using skeleton features, and it surpasses existing methods.

The key contributions of our works can abridge as follows. First, we propose a novel method for noise-robust skeleton-based action recognition, called Predictively Encoded Graph Convolutional Network (PeGCN), which performs better than the existing state-of-the-art methods on either general skeleton-based action recognition and that with noisy skeleton samples. Second, predictive encoding loss on latent space captures useful representations to predict complete skeleton features from noisy skeleton and improves action recognition performance with noisy samples. In addition to these contributions, we also provide comprehensive experiments on skeleton-based action recognition with noised samples. Our experiments include various ablation studies and comparisons with existing GCN based methods. 

\section{Skeleton-based action recognition using deep learning}
\label{sec:2}
The recent success of deep-learning techniques had a significant impact on the studies for human action recognition. To model spatio-temporal features of actions, many works \cite{liu2016spatio,si2018skeleton,lee2017ensemble,zhang2017view,zhu2016co,shahroudy2016ntu} attempt to extract appearance information with Convolutional Neural Networks (CNNs) and temporal information with Recurrent Neural Networks (RNNs). TS-LSTM \cite{lee2017ensemble} uses multiple temporal-windows to handle both short/mid/long-range actions dynamically. Zhang \etal \cite{zhang2017view} proposed view-adaptive action model (VA-LSTM) which is robust to view point change. However, CNN/LSTM based methods usually represent the skeleton data as a sequence of vectors which cannot express the dependencies enough between related joints. The skeleton model can be seen as a graph structure where joints and bones correspond to the vertices and edges, respectively. 

Recently, ST-GCN \cite{yan2018spatial} successfully adopted graph convolution networks (GCNs) to handle graphs in arbitrary forms and it was the first method which applied GCNs to the skeleton-based action recognition. After Yan \etal \cite{yan2018spatial} proposed ST-GCN, lots of works using graph convolution networks (GCNs) were proposed. The GCNs have two main approaches to apply: spectral approach \cite{li2018spatio} and spatial approach \cite{shi2019two,shi2019skeleton,li2019actional,si2019attention}. The spectral approach first performs Eigen decomposition on graph Laplacian matrix to get Eigenvalues and Eigenvectors. With these Eigen features, graph convolution is performed on sub-graph with graph Fourier transform. In this way, no locally-connected node partitioning is required. On the other hand, in spatial perspective method performs graph convolution directly on graph nodes with it's neighbourhood nodes. This approach is widely accepted in action recognition (\eg ST-GCN) since it takes less computational cost than spectral approach. 

The main drawbacks of ST-GCN is the spatial graph which is predefined only relying on the physical structure of human body and is fixed to all the GCN layers. These drawbacks make hard to capture not only the relationships between closely related joints such as both two hands in hand-related action, but also the dynamics of each action. To tackle these limitations, many methods \cite{shi2019two,shi2019skeleton,song2019richly,li2019actional,si2019attention} were proposed to build adaptive graph to pay attention dynamically to each joint based on the performing action.  The adaptive graph is trainable mask which learns relationships between any joints which can increase both flexibility and generality in constructing the graph. Shi \etal \cite{shi2019two} proposed 2s-AGCN model which has two adaptive graphs: 1) global-graph and 2) local-graph. Both of them are trained and updated jointly with CNNs in end-to-end manner. The global-graph learns common patterns for all the samples while local graph learns unique patterns of each individual sample. Lie \etal \cite{li2019actional} proposed actional links (A-links) to learn action-specific dependencies, and structural-links (S-links) for higher-order relationships between joints. 

While most of works were using undirected graph, Shi \etal \cite{shi2019multi} proposed directed graph based model (DGNN) where direction of the graph plays important role in graph convolution for updating features of edges and vertices. Si \etal \cite{si2019attention} combines LSTM with GCN (AGC-LSTM ) to learn a spatio-temporal representations from sequential skeletons, while most of GCN based action recognition models acquire temporal information with 1d-convolution on the temporal-axis. Spatial based GCNs usually distribute graphs into multiple sub-graphs with distance partitioning or spatial configuration partitioning proposed in \cite{yan2018spatial}. In contrast to these common partitioning strategies, Thakkar \etal \cite{thakkar2018part} proposed part-based GCN (PB-GCN) that learns relationships between five body parts.

\section{Predictively Encoded Graph Convolutional Networks}
\label{sec:3}
\subsection{Motivation and Intuition}
\begin{figure*}[t]
\centering
\includegraphics[width=\textwidth,height=4.5cm]{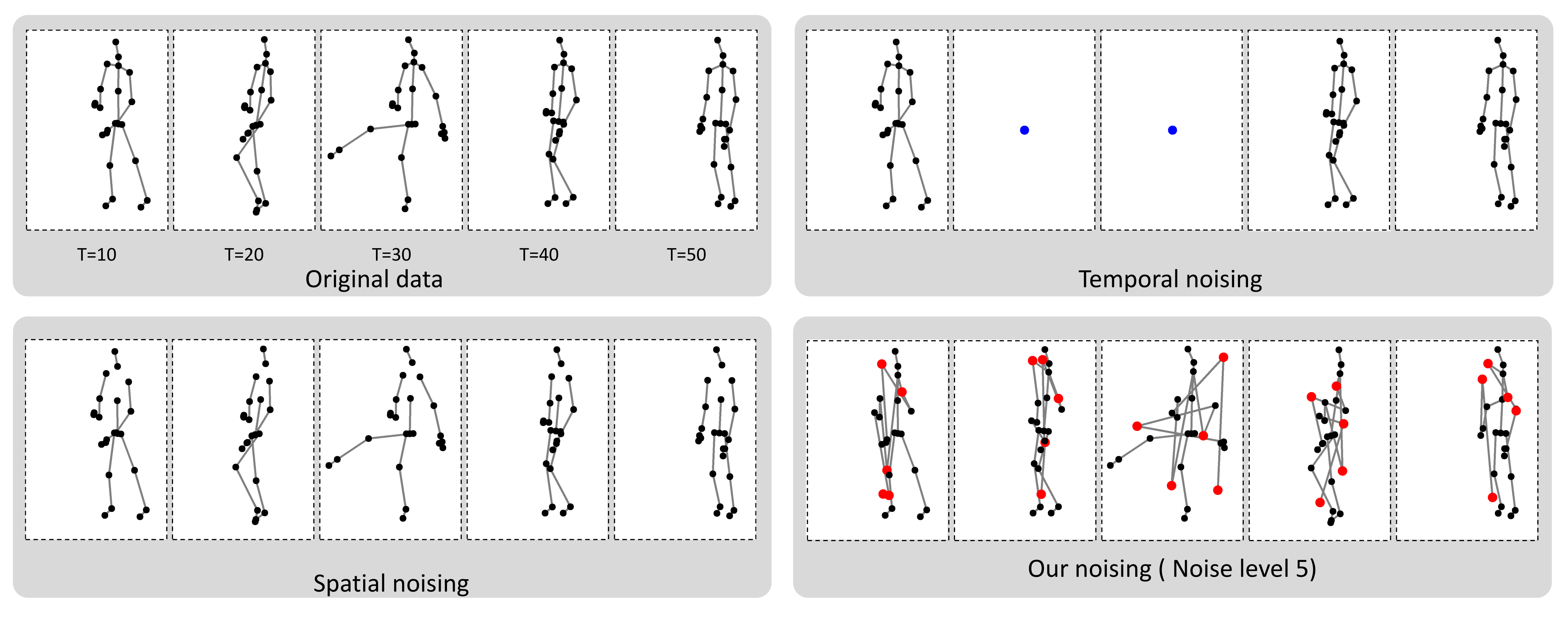}
\caption{Illustrations of various types of noisy skeletons. $T$ is the frame order associated with each skeleton.
 (a) is original skeleton samples. (b) and (c) are the skeleton samples considered by Song \etal \cite{song2019richly}, which are spatially and temporally occluded. (c) is the noisy skeleton sample generated by our noising approach using a noise level 5.}
\label{fig:noise_typ}
\vspace{-4ex}
\end{figure*}

For developing the precise action recognition method, it is important to learn a global representation which can represent every detail of given video clip for the entire time period. To learn a suitable global representation, a model needs outstanding generalization ability which can be robust to the diverse types of noise. Variation of skeleton features depending on geometric conditions, such as a viewpoint of cameras or acting objects, can be regarded as a sort of noise skeleton features. Missing of skeleton features (a.k.a., incomplete skeleton features \cite{song2019richly} (see Fig. \ref{fig:noise_typ}) by spatial or temporal occlusions, are also a kind of noisy skeleton features. These noise patterns are inherently unpredictable. It is, therefore, intractable to model noise information explicitly in a data-driven approaches.

Deep learning is well known as an effective way to improve generalization performance of a model for various visual recognition studies \cite{krizhevsky2012imagenet,du2015hierarchical,badrinarayanan2017segnet,girshick2015fast}. GCN is a unified framework of a graph structure and deep learning, so it also has advantage improved generalization performance. Based on this advantage, The dominant approach to training the skeleton-based action recognition methods based on GCNs is initially extracting information from skeleton samples using GCNs and then computing the unimodal loss \eg cross-entropy \cite{yan2018spatial,shi2019two,shi2019skeleton,song2019richly,li2019actional,si2019attention}. It can be regarded as direct end-to-end learning for a model $p(\bar{o}|x)$ between skeleton samples $x$ and a corresponding acting classes $\bar{o}$. However this approach, which directly derives a mapping model for $p(\bar{o}|x)$ and $p(\bar{o}|x')$ from a complete sample $x$ or an incomplete sample $x'$ to class label $\bar{o}$, is computationally intensive and a waste of representation capacity of the model. For example, the mapping between $x$ and $\bar{o}$ directly can be thought as using every detail of input samples all the time whether it is necessary or not. A slight noise, which can be alleviated during generalization via a non-linear network structure, does not need to be considered seriously. As a result, it may not suitable to derive a mapping model $p(\bar{o}|x)$ directly for deriving the optimal global representation.

The key insight of PeGCN for noise-robust skeleton-based action recognition is to learn the representations that encode to the underlying shared information between complete sample and noisy sample via predicting missing information in the latent space. This idea is inspired by the predictive coding \cite{1055126,atal1970adaptive,oord2018representation} which is one of the oldest techniques in signal processing for data compression, and recently it is applied to unsupervised learning for learning word representations \cite{mikolov2013efficient} by predicting neighbouring words. The approach to latent space has the following advantages: First, since action recognition processes relatively long time samples than the others including event detection \cite{yu2018joint,8580568} or change detection \cite{hussain2013change}, action recognition models need to infer more global structure. When inferring the global structure, high-level information \ie latent space, is more suitable than the low-level information. Second, the global noise on the latent representation is likely to be a serious noise which can affect the recognition performance seriously than local noise which can be reduced via non-linear weighted kernel structures of deep learning.

\begin{figure*}[t]
	\vspace{-0.2cm}
	\begin{center}
		\centerline{\includegraphics[width=\linewidth, height=4cm]{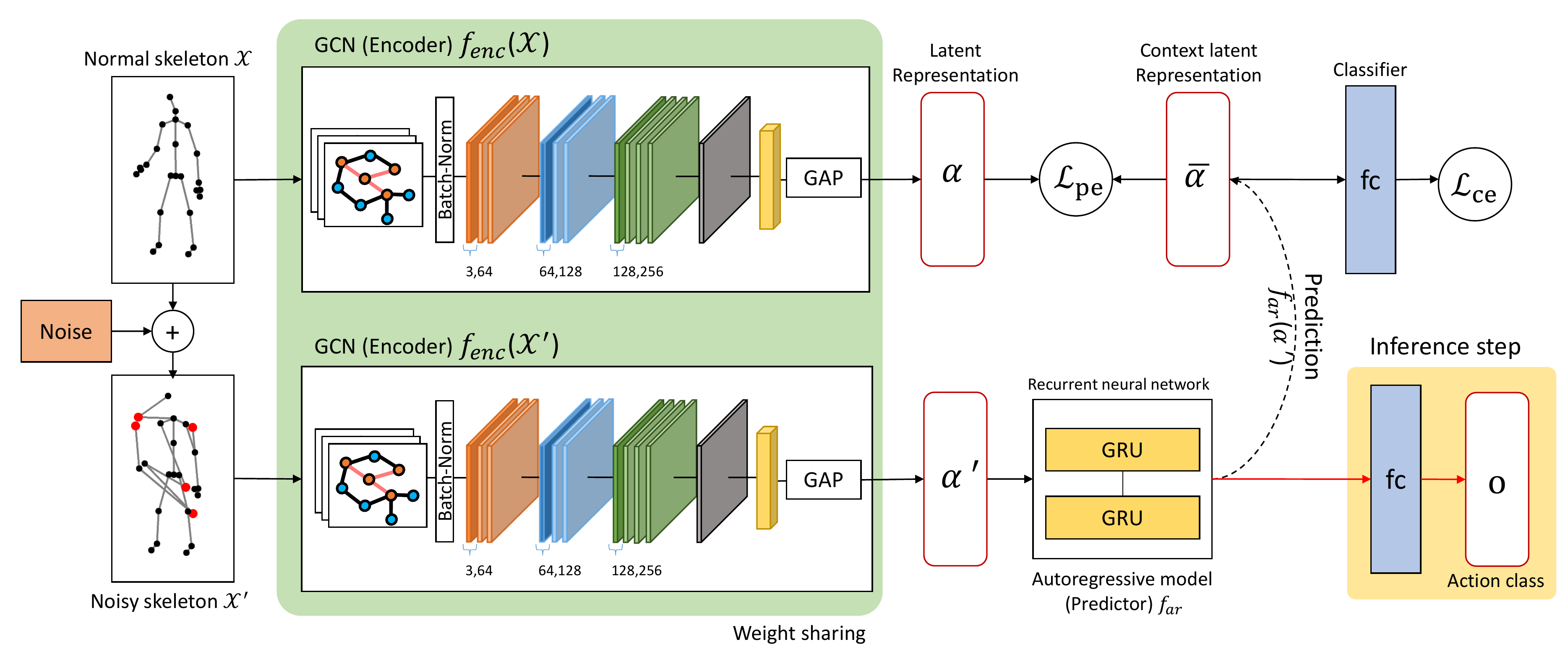}}
		\caption{The pipeline of PeGCN for the training and inference steps. The backbone network is the GCN of Js-AGCN \cite{shi2019two}. The figures under each layer are the dimensionalities of input and output channels, respectively and GAP is global average pooling operation. The black solid and dotted lines denote the pipelines for the training step. The red solid lines denote the extra pipelines for the inference step.}
		\label{fig:1}
	\end{center}
	\vspace{-8ex}
\end{figure*}

When predicting proper information from noise skeleton features, we initially map the normal skeleton features $x$ and noise skeleton feature $x'$ into compact distributed vector representations (a.k.a., latent features) $\alpha$ and $\alpha'$ respectively, via non-linear mapping function, and train the model in a way that maximally preserves the mutual information between $\alpha$ and $\alpha'$. The mutual information is defined by, 
\begin{equation}
I(\alpha;\alpha') = \sum_{\alpha,\alpha'}p(\alpha,\alpha')\text{log}\frac{p(\alpha|\alpha')}{p(\alpha)}.
 \label{eq:mutual_info}
\end{equation}
By maximizing the mutual information between two encoded representations (which are bounded by the MI between the input signals), we extract the underlying latent variable robust to the global noise. 

\subsection{Structural details}
Fig. \ref{fig:1} illustrates the pipeline of PeGCN on the training and inference. PeGCN consists of a GCN module $f_{\text{gcn}}$ and an autoregressive module $f_{\text{ar}}$. The GCN module $f_{\text{gcn}}$ encodes skeleton samples into a latent space $\alpha* = f_{\text{gcn}}(x*)$, where $*$ indicate the input types: normal one $x$ and $\alpha$ or noise one $x'$ and $\alpha'$. The autoregressive module $f_{\text{ar}}$ summarizes the latent representation and produces a context latent representation $\bar{\alpha}=f_{\text{ar}}(\alpha*)$, where $\alpha*$ can be defined by $\alpha$ and $\alpha'$ depending on the corresponding input skeletons. 

In the training step, the normal skeleton samples $x$ and the corresponding noisy skeleton samples $x'$ are provided. First, the GCN module $f_{\text{gcn}}$ produces latent representations $\alpha$ and $\alpha'$ from $x$ and $x'$, respectively. Next, the autoregressive module $f_{\text{ar}}$ extracts the context latent representation $\bar{\alpha}$ from the latent representation $\alpha'$ only. As argued in the previous section, we do not train a model by directly deriving $p(o|x')$ or $p(o|\alpha')$. Instead, PeGCN is trained in the way to maximize the mutual information (Eq. \ref{eq:mutual_info}) between the two latent representations, $\alpha$ and $\alpha'$ of the normal and noisy skeleton samples, by modeling a density ratio which preserves the mutual information (Eq. \ref{eq:mutual_info}) between $\alpha$ and $\alpha'$ as follows:
\begin{equation}
I(\alpha;f_{\text{ar}}(\alpha')) = \sum_{\alpha,f_{\text{ar}}(\alpha')}p(\alpha,f_{\text{ar}}(\alpha'))\text{log}\frac{p(\alpha|f_{\text{ar}}(\alpha'))}{p(\alpha)}.
\label{eq:ml_autoregressive}
\end{equation}
By using $I(\alpha;\bar{\alpha})$ with autoregressive module $f_{\text{ar}}$, we relieve the model from modelling the high dimensional distribution $x$ or $x'$. Although we cannot evaluate $p(x)$ or $p(\bar{o}|x)$ directly, we can use samples from these distributions, allowing us to use the technique as Noise-Contrastive Estimation \cite{gutmann2010noise,mnih2012fast,jozefowicz2016exploring} and Important Sampling \cite{bengio2008adaptive}. The output of autoregressive module $\bar{\alpha}$ can be used if extra context from the representation is useful. One such example is speech recognition, that the receptive field of $\alpha$ may not contain enough information to capture phonetic content. In other cases, where no additional context is required, $\alpha$ might be better instead. 

\begin{figure*}[t]
        \centering
        \begin{subfigure}{0.23\textwidth}
            \includegraphics[width=\textwidth,height=2.5cm]{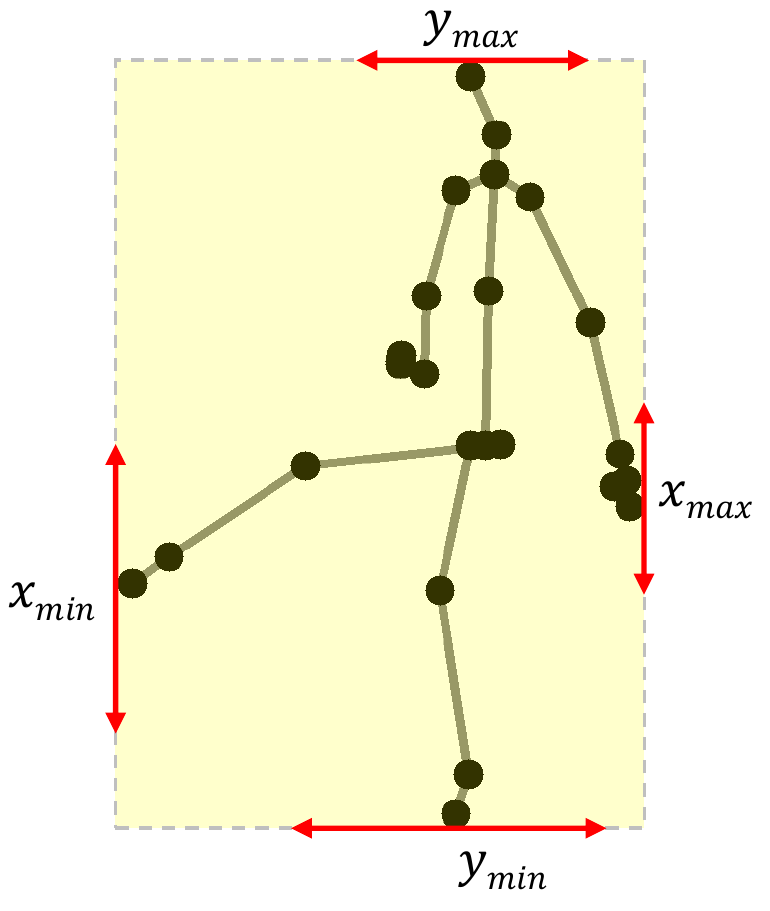}
	\caption{}
        \end{subfigure}
        \hfill
        \begin{subfigure}{0.73\textwidth}
            \includegraphics[width=\textwidth,height=2.5cm]{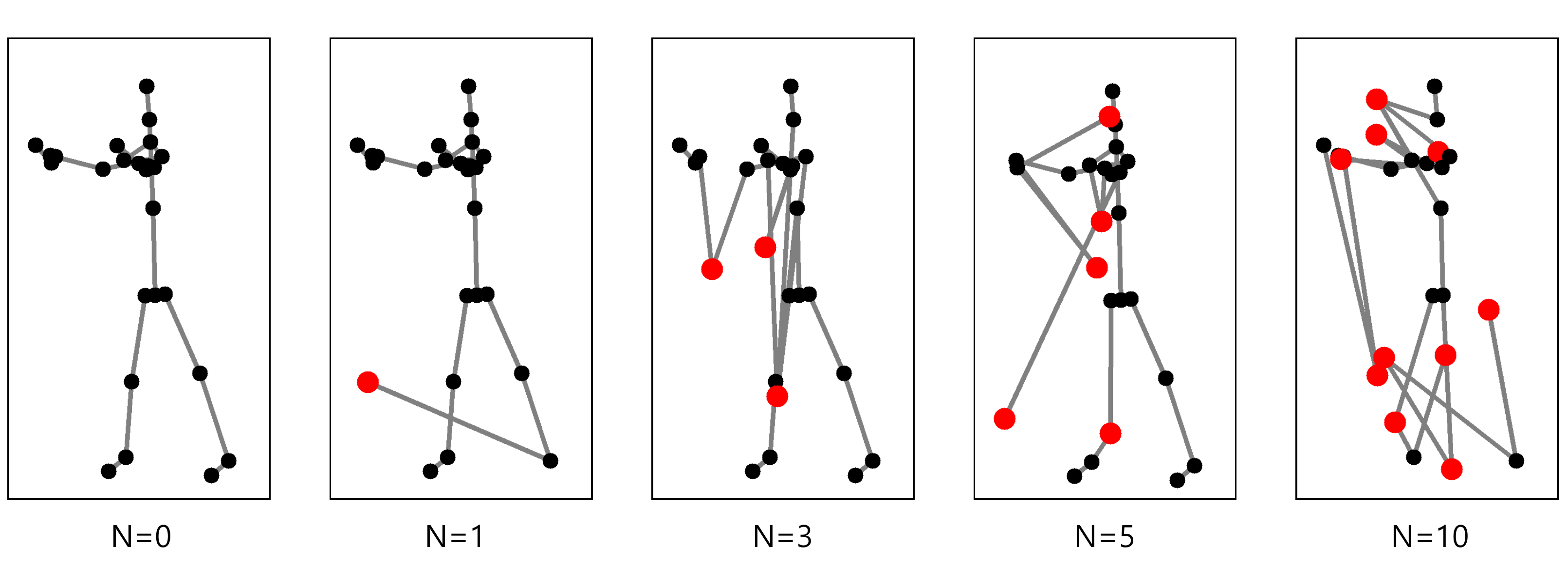}
	\caption{}
        \end{subfigure}
        \hfill
        \caption{Illustrations of how to set the candidate scope for generating noise joint and the examples of noise skeleton samples depending on the noise level. (a) illustrates that how to define the scope for generating noise joint using a given skeleton sample. (b) shows the noise skeleton samples created from original sample depending on the noise level.}
\label{fig:gen_noise}  
\vspace{-3ex}
\end{figure*}

The noise skeleton features $x'$ are generated by adding some noise to randomly picked joints in the original skeleton samples $x$. The noise is generated based on the bounding box computed using the minimum and maximum values of the x, y, and z coordinates of skeleton samples (Fig. \ref{fig:gen_noise}(a)). When generating the noise samples in the training and test steps, we set the noise level which is the parameter to decide how many joints would be noised. The generated noisy samples depending on the noise level are represented in Fig. \ref{fig:gen_noise}(b). The noise skeleton samples that we are regarding in this paper, are different from the spatially or temporally occluded skeleton samples that considered in Song \etal \cite{song2019richly} (see Fig. \ref{fig:noise_typ}). In the real scenarios, the missed joints in the occluded skeleton samples can be defined by a set of joints that have low likelihoods or confidences than a pre-defined threshold. However, a noise is inherently unpredictable so that assumption may not practical. 

The backbone network for our GCN module $f_{\text{gcn}}$ is the GCN part of of Js-AGCN \cite{shi2019two}, which composed of adaptive graph convolutional layers, which can make the topology of the graph optimized together with the other parameters of the network in end-to-end learning manner. The adaptive convolutional layer is defined by,
\begin{equation}
\boldsymbol{f}_{\text{out}}=\sum^{K_{v}}_{k}\boldsymbol{W}_{k}\boldsymbol{f}_{\text{in}}(\boldsymbol{A}_{k}+\boldsymbol{B}_{k}+\boldsymbol{C}_{k}),
 \label{eq:2sagcn}
\end{equation}
where $\boldsymbol{A}_{k}$ is the original normalized adjacency matrix for GCN, $\boldsymbol{B}_{k}$ is the trainable matrix for global attention and $\boldsymbol{C}_{k}$ is a data-dependent graph for learning a unique graph for each sample. We employ the GCN of 2s-AGCN without the fully connected networks located on the after the GCN. 

We use RNNs using GRUs \cite{chung2014empirical} for the autoregressive module $f_{\text{ar}}$. This can be easily replaced by other linear transformation or non-linear networks. The detilas of dimensionalities of the GCN module and the autoregressive module of PeGCNs are represented in Appendix \ref{apx:1}. Note that any type of GCN model and autoregressive model can be applied in the proposed method. Probably, more recent advancements in GCNs and autoregressive modelling could help improve results further. 

\subsection{Training and inference}
Both the GCN and autoregressive modules are jointly trained to optimize the loss in order to maximize the mutual information between two latent representations of normal and noise skeleton features, which we call predictive encoding loss. With given set for normal skeleton samples $\boldsymbol{x}\in\{x_{i}\}_{i=1:n}$ of $n$ samples and the corresponding noise skeleton samples $\boldsymbol{x'}\in\{x'_{i}\}_{i=1:n}$, the predictive encoding loss is defined by,
\begin{equation}
\mathcal{L}_{\text{pe}} = - \mathop{{}\mathbb{E}}_{X,X'}\left[\log \frac{p(f_{\text{gcn}}(x),f_{\text{ar}}(f_{\text{gcn}}(x')))}{\sum_{x \in X} p(f_{\text{gcn}}(x))}\right].
 \label{eq:pe_loss}
\end{equation}
Optimizing this loss will result in $I(\alpha,\alpha')$ estimating the density ratio in Eq. \ref{eq:mutual_info}. It is theoretically and experimentally demonstrated by Ooord \etal \cite{oord2018representation}.

Action recognition should identify an action class of given skeleton sample. Using  $\mathcal{L}_{\text{pe}}$ only can not achieve this goal since it is only focused on the maximizing mutual information between two latent representation. Therefore, as similar to other studies \cite{shi2019two,song2019richly,shi2019multi}, the cross-entropy loss is exploited as follows, 
\begin{equation}
\begin{split}
   \mathcal{L}_{ce}=-\sum_{i}^{C}\bar{o}_{i}\text{log}(o_{i}),
   \end{split}
 \label{eq:ce_loss}
\end{equation}
where $C$ is the numbers of action classes. $\bar{o}$ is a given annotation for an action sample, and $o$ is the output of the fully connected network for classification task on the inference step. 

Consequently, to train the noise-robust skeleton-based action recognition model, the total loss functions is straightforwardly defined by the sum of the cross-entropy loss $\mathcal{L}_{\text{ce}}$ ,and the proposed predictive loss function with the balancing weight $\lambda$. It is represented as follows:
\begin{equation}
\begin{split}
   \mathcal{L}_{\text{total}}=\mathcal{L}_{\text{ce}}+\lambda\mathcal{L}_{\text{pe}}.
   \end{split}
 \label{eq:total_loss}
\end{equation}
In all our experiments, $\lambda$ is set by 0.1 for the best performance. 

The action recognition using PeGCN is straightforward. In the test step, the GCN module $f_{\text{gcn}}$ encodes an input skeleton sample into the latent space, and the autoregressive model $f_{\text{ar}}$ summarizes the latent feature and generate the context latent representation $\bar{\alpha}$. The $\bar{\alpha}$ is used as an input of a fully connected networks for action recognition (Fig. \ref{fig:1}).

\section{Experiments}
\label{sec:4}
\subsection{Experimental setting}
To evaluate the action recognition performances of PeGCN and other methods on noise skeleton samples, we use NTU-RGB+D dataset \cite{shahroudy2016ntu}, which is one of the largest datasets in skeleton-based action recognition, and Kinetics-skeleton (a.k.a., Kinetics) dataset generated from the Kinetics dataset \cite{kay2017kinetics} containing 34,000 video clips. Two experimental protocols: 1) Cross-view (CV) and 2) Cross-subject (CS) are applied for the experiments using NTU-RGB+D dataset. The detail explanations of the two datasets are described in Appendix \ref{apx:2}.

The settings of common hyperparameters to train PeGCNs are as follows. The numbers of epochs are 50 and 65 for NTU-RGB+D dataset and Kinetics-skeleton dataset, respectively. Since our computational resources are limited, the batch size reduced to 32 and it is the half of original batch size of our backbone network \cite{shi2019two} which can affect the action recognition performance of PeGCNs negatively. Stochastic gradient descent and the weight decay are utilized as optimization algorithms. The source code of PeGCN is publicly available on \url{https://github.com/andreYoo/PeGCNs.git}. The source code includes the feed function to generate the noise skeleton samples. The experiments are categorized into two parts. One is for the ablation study, and another is for the comparison with existing state-of-the-art methods.

\subsection{Ablation study}
\textbf{Experimental protocol} We have conducted the performance analysis depending on the hyperparameter settings of PeGCN. The hyperparameters that significantly affect the action recognition performance of PeGCNs are the noise level and the composition of loss functions. The performance analysis depending on the setting of noise level and the composition of loss functions in the training step is as follows. First, we construct two PeGCN models trained by $\mathcal{L}_{\text{ce}}$ (PeGCN$_{\text{cd}}$) and $\mathcal{L}_{\text{total}}$ (PeGCN$_{\text{total}}$), and each model is trained with 1, 3, and 5 noise-levels. Other parameters are set as exactly same to the parameter setting mentioned in the above section. Next, we evaluate these models with noise level between 0 to 5. We have observed the trends of the cross-entropy losses and the predictive coding losses of these models and compared the action recognition accuracies. For efficient experiments, the ablation study is only conducted with the CV protocol of NTU-RGB+D dataset. 

\begin{figure*}[t]
    \centering
\small
\begin{minipage}{.9\linewidth}
    \begin{subfigure}{.49\textwidth}
        \includegraphics[width=\textwidth,height=3cm]{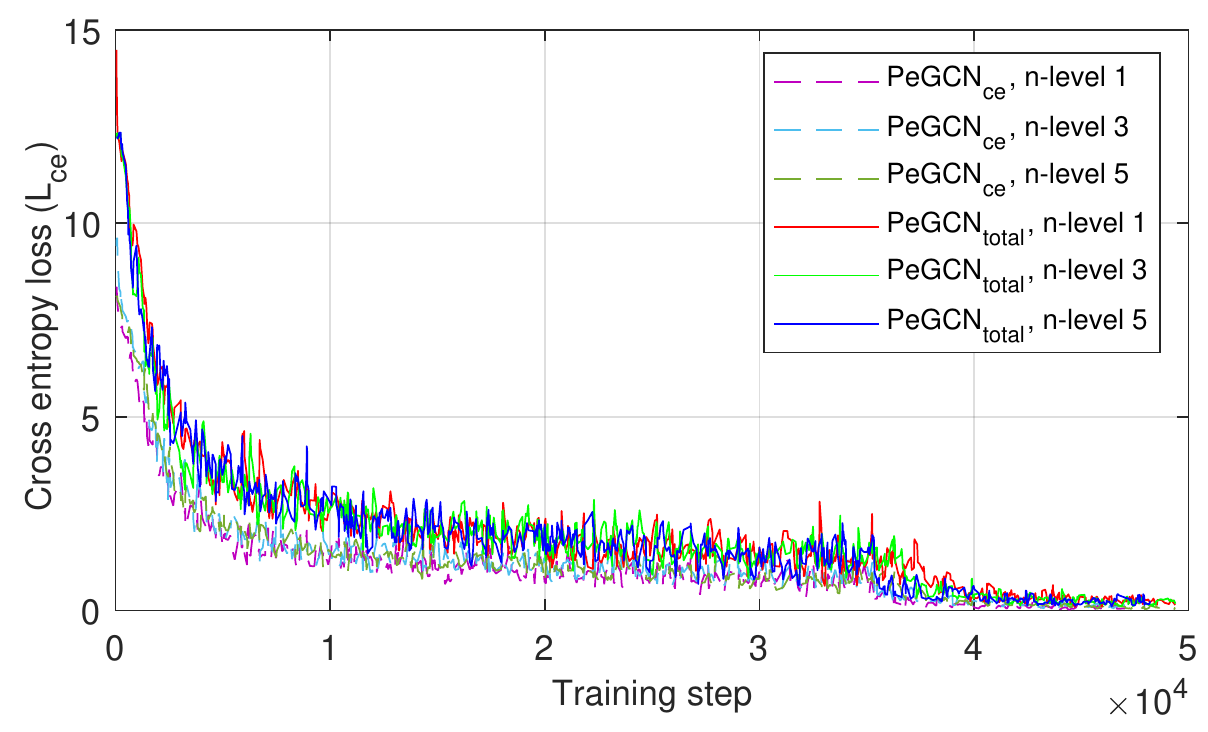}
    \caption{}
    \end{subfigure}
    \begin{subfigure}{.49\textwidth}
        \includegraphics[width=\textwidth,height=3cm]{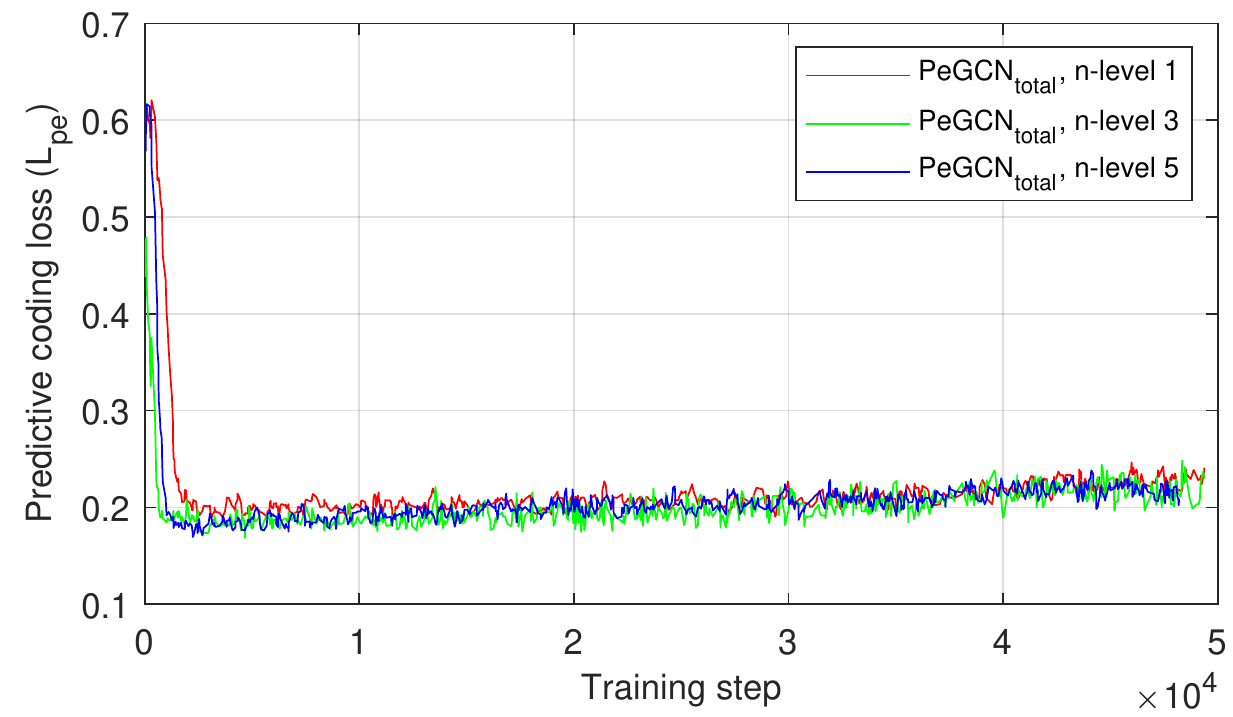}
    \caption{}
    \end{subfigure}
    \hfill
    \small
    \end{minipage}
    \caption{\small Trends of the cross-entropy and predictive encoding losses according to the noise level in training PeGCNs on the CV protocol of NTU-RGB+D dataset. (a) indicates the curves of the cross-entropy functions $\mathcal{L}_{\text{ce}}$. (b) represents the curves of the predictive encoding losses $\mathcal{L}_{\text{pe}}$. The curves of $\mathcal{L}_{\text{pe}}$ are constructed from the PeGCNs trained by $\mathcal{L}_{\text{total}}$.}
     \label{fig:ce_pe_graphs} 
\begin{minipage}{\linewidth}
\centering
     \resizebox{\textwidth}{!}{
    \begin{tabular}{c c c c c c c c cc c}
		\toprule
		\midrule
		 \multirow{2}[4]{*}{Train N-level} & \multicolumn{9}{c}{Test N-level} \\ 
		\cmidrule(rl){2-11}
		 & \multicolumn{2}{c}{1} & \multicolumn{2}{c}{3}  & \multicolumn{2}{c}{5}  & \multicolumn{2}{c}{7} & \multicolumn{2}{c}{10} \\
			\cmidrule(rl){2-11}
		 & \quad Top 1\quad  &\quad Top 5\quad &\quad Top 1\quad  &\quad Top 5\quad &\quad Top 1\quad  &\quad Top 5\quad &\quad Top 1\quad  &\quad Top 5\quad &\quad Top 1\quad  &\quad Top 5\quad \\
		 \midrule
		\multicolumn{11}{c}{Training without the predictive encoding loss $\mathcal{L}_{\text{pe}}$ (PeGCN$_{\text{ce}}$)} \\ 
		\midrule
		  1 &  90.31($\pm$0.14)&	93.42($\pm$0.07)&	79.51($\pm$0.09)&	91.65($\pm$0.07)&	67.24($\pm$0.13)&	72.42($\pm$0.03)&	61.08($\pm$0.06)&	79.37($\pm$0.03) &	55.39($\pm$0.11)&   70.42($\pm$0.06)\\
		  3 &  91.96($\pm$0.09)&	95.97($\pm$0.08)&	83.52($\pm$0.21)&	93.91($\pm$0.10)&	76.31($\pm$0.15)&	89.81($\pm$0.07)&	71.50($\pm$0.09)&	80.31($\pm$0.05) &	64.12($\pm$0.17)&	73.42($\pm$0.04)\\
		  5 &  92.87($\pm$0.08)&	97.25($\pm$0.06)&	91.62($\pm$0.14)&	95.42($\pm$0.08)&	87.43($\pm$0.11)&	90.31($\pm$0.06)&	83.26($\pm$0.09)&	88.42($\pm$0.6) &	74.37($\pm$0.16)&	81.26($\pm$0.09)\\
		\midrule
		\multicolumn{11}{c}{Training with the predictive encoding loss $\mathcal{L}_{\text{pe}}$ (PeGCN$_{\text{total}}$)} \\ 
		\midrule
		  1 &  91.72($\pm$0.05)&	98.31($\pm$0.03)&	86.42($\pm$0.08)&	95.52($\pm$0.05)&	84.52($\pm$0.10)&	91.61($\pm$0.01)&	79.41($\pm$0.07)&	84.52($\pm$0.01) &	61.25($\pm$0.13)&	74.36($\pm$0.07)\\
		  3 &  92.63($\pm$0.05)&	98.98($\pm$0.03)&	91.92($\pm$0.09)&	97.51($\pm$0.03)&	89.52($\pm$0.09)&	94.12($\pm$0.04)&	81.25($\pm$0.09)&	91.52($\pm$0.05) &	78.12($\pm$0.12)&	90.21($\pm$0.07)\\
		  5 & \textbf{93.21}($\pm$0.04)& \textbf{98.97}($\pm$0.02)&	\textbf{92.78}($\pm$0.09)&	\textbf{98.91}($\pm$0.04)&	\textbf{92.24}($\pm$0.08)&	\textbf{98.81}($\pm$0.03)&	\textbf{91.08}($\pm$0.06)&	\textbf{98.52}($\pm$0.03) &	\textbf{89.39}($\pm$0.11)&	\textbf{98.29}($\pm$0.06)  \\
		\midrule
		\bottomrule
	\end{tabular}
	}
\end{minipage}
\captionof{table}{\small Action recognition accuracies of PeGCNs depending on the setting of the loss functions and the noise level in the training step on the CV protocol of NTU-RGB+D dataset. The figures in parentheses are standard deviations. The boldface figures denote the highest performance for each experiment.}
\label{tbl:abl} 
\vspace{-4ex}
\end{figure*}

\textbf{Experimental results}. Table \ref{tbl:abl} shows that action recognition accuracy depending on the noise levels and the setting of the loss functions in the training step. The best accuracy is achieved by the PeGCN$_{\text{total}}$ trained with noise level 5. Its achieves 93.21 of accuracy in noise level 1 and 89.39 of accuracy for the noise level 10  in the test step, respectively. The PeGCN$_{\text{ce}}$ trained with the noise level 5, achieves 92.87 of accuracy in noise level 1. The PeGCN$_{\text{total}}$ trained with noise-level 5, achieves 92.24 of accuracy in the evaluations with the noise level 5. It also produces 89.39 of accuracy in the test with the noise level 10. On the other hand, the PeGCN$_{\text{ce }}$ trained with noise-level 5, obtains 87.43 of accuracy on the test with noise level 5. The quantitative results demonstrate that if models are trained at the same noise level, the model trained with the total loss function $\mathcal{L}_{\text{total}}$ usually performs better, and it also suggests that the performance degradation of the PeGCNs trained by the cross-entropy loss only, is much faster than the others.  Not only quantitative results, but also the trend of each losses show the efficiency of the predictive encoding loss when learning the noise-robust representation. The trends of cross-entropy losses in the ablation studies (see Fig \ref{fig:ce_pe_graphs}(a)) show that the curves of the PeGCNs trained by the cross-entropy losses only, are converged faster than the PeGCNs trained by the total loss $\mathcal{L}_{\text{total}}$ usually. It can be thought that the PeGCN trained with the cross-entropy loss only is easier to converged into the poor locally optimized solution than the others.

Interestingly, The trend of predictive encoding loss (see Fig \ref{fig:ce_pe_graphs}(b)) shows that the curve of the PeGCN$_{\text{total}}$ trained by noise level 5 is relatively lower than that of the PeGCN$_{\text{total}}$ trained by noise level 1 or 3. In the graphical comparison, PeGCNs trained by $\mathcal{L}_{\text{total}}$ and  $\mathcal{L}_{\text{ce}}$ are compared to each other. These trends can be interpreted as a difficulty of learning with highly noised samples. In the training step, a higher noise level can provide more diversity in the training samples than the lower noise level. Consequently, the ablation study demonstrates that the higher noise-level in the training step can improve the action recognition performance in the test step, but it is not linearly proportional. For the efficient experiments, further studies for comparing PeGCN with existing state-of-the-art methods are only conducted with PeGCN$_{\text{total}}$ trained with noise level 5. 

\subsection{Comparison with existing state-of-the-art methods}
\textbf{Experimental protocol}. Our experiments include either the experiments with normal skeleton samples or that with noisy samples. Basically, we follow the general experimental protocol described in NTU-RGB+D dataset \cite{shahroudy2016ntu} and Kinetics-skeleton dataset \cite{kay2017kinetics}. For both datasets, top-1 and top-5 accuracies are computed for the performance comparison. In the experiments using NTU-RBGD dataset, both CV and CS protocols are applied. To evaluate action recognition performance on the noisy setting, we artificially generate noisy samples as follows: First, the number of joints (a.k.a. noise level) for assigning noise values is determined manually. Second, according to the noise level, the joints which would be assigned by noise value, are randomly picked.  The selected joints are constant for all frames in the video clip. 

After which joints will be noised is decide, random values generated from the bounding-box are assigned to each selected joint in every frame (see Fig. \ref{fig:gen_noise}(a)). To reduce the volatility of performance due to the randomness of noised joints, all experiments are iteratively conducted for 10 times, and the average and standard deviation for the results are used for the comparison. The examples of the artificially generated noisy skeleton samples are illustrated in Appendix \ref{apx:3}.

Predominantly, we have tried to compare PeGCN with recently proposed state-of-the-art methods. For efficient experiment and fair comparison, methods which were proposed before 2018 or performances are lower than ours by 5\% in normal skeleton evaluation (\eg \cite{fernando2015modeling,du2015hierarchical,shahroudy2016ntu,liu2016spatio,zhang2017view}) are excluded for the comparison (see Table \ref{tab:normal}). Particularly, in the experiments using noisy skeleton samples, methods, which source code did not be released by paper authors, are excluded in the experiments \cite{shi2019skeleton,peng2019learning}. Even if source code exists, some methods are excluded by the following criteria: First, the source codes have released from non-authors \cite{shi2019skeleton}. Second, the paper is not officially published yet on a journal or a conference \cite{peng2019learning}. Third, the source codes are argued by other people that they can not obtain the performance reported on a paper \cite{shi2019skeleton}. The detail information of the source code and the pre-trained weight for each model are described in Appendix \ref{apx:4}. 

\textbf{Experiment with normal skeletons.} Initially, we compare PeGCN with other existing state-of-the-art methods on normal skeleton samples. For the consistency of the experiments, several methods are tested using publicly available source codes by ourselves \cite{yan2018spatial,song2019richly,thakkar2018part,shi2019two}. Table \ref{tab:normal} contains the top 1 accuracies on the CS and CV protocols of NTU-RGB+D dataset and the top 1 and top 5 accuracies on Kinetics dataset. In the experiments, PeGCN$_{\text{total}}$ achieves 85.6 and 93.4 accuracies on the CS and CV protocols of NTU-RGB+D dataset, respectively. PeGCN$_{\text{total}}$ produces 34.8 and 57.2 for Top 1 and top5 accuracies in Kineitcs-skeleton dataset. The state-of-the-art performance is achieved by MS-AAGCN \cite{shi2019multi} with 90.0 for CS protocol and 96.2 for CV protocol. The second highest performance is achieved by DGNN \cite{shi2019skeleton} which recorded 89.9 and 96.1 for CS and CV protocol, respectively. In Kinetics-skeleton dataset, MS-AAGCN \cite{shi2019multi} scores 37.8 for top-1 and 61.0 for top-5. MS-AAGCN scores the second-highest performance again with 37.8 and 61.0 for top-1 and top-5, respectively.

\begin{wraptable}{r}{60mm}
\vspace{-4ex}
	\centering
	\resizebox{0.5\textwidth}{!}{
	\begin{tabular}{l c c c c c }
		\toprule
		\midrule
		\multirow{2}[2]{*}{Methods}  & 	\multirow{2}[2]{*}{Nets}  & \multicolumn{1}{c}{CS} & \multicolumn{1}{c}{CV}& \multicolumn{2}{c}{Kinetics}  
		\\ \\ \cmidrule(rl){3-3}\cmidrule(rl){4-4}\cmidrule(rl){5-6}
		&  & T1 & T1 & T1 & T5   \\
		\cmidrule(l){1-6}
		\multicolumn{1}{l}{VA-LSTM \cite{zhang2017view}}      & LSTM & 80.7 & 88.8  &- &- \\
		\multicolumn{1}{l}{Clips+CNN+MTLN \cite{ke2017new}} &  CNN & 79.6 & 84.8 &  - &- \\
		\multicolumn{1}{l}{Synthesized CNN \cite{liu2017enhanced}} & CNN &  80.0  & 87.2 &- &- \\
		\multicolumn{1}{l}{3scale ResNet152 \cite{li2017skeleton}} & CNN & 85.0 &  92.3   &- &- \\
 	    \midrule
 		\multicolumn{1}{l}{DPRL+GCNN \cite{tang2018deep}} & GCN &  83.6 & 89.8  &-   &- \\
	    \multicolumn{1}{l}{AGC-LSTM \cite{si2019attention}} &GCN+LSTM &89.2  &95.0 &-&-  \\
		\multicolumn{1}{l}{AS-GCN \cite{li2019actional}} &GCN &   86.8   &  94.2 & 34.8 & 56.5   \\
		\multicolumn{1}{l}{ST-GCN$^{*}$ \cite{yan2018spatial}} & GCN & 81.6  & 88.8 & 31.6 & 53.7 \\
		\multicolumn{1}{l}{2s RA-GCN$^{*}$ \cite{song2019richly}}& GCN &  85.8 & 93.0 & - &- \\
		\multicolumn{1}{l}{3s RA-GCN$^{*}$ \cite{song2019richly}}& GCN &  85.9 & 93.5  & - &- \\
        \multicolumn{1}{l}{PB-GCN$^{*}$ \cite{thakkar2018part}} & GCN  &  87.0 & 93.4  & - & -\\
        \multicolumn{1}{l}{Js-AGCN$^{*}$ (Backbone) \cite{shi2019two}} &GCN   & 85.4 & 93.1  & 34.4 & 57.0 \\
		\multicolumn{1}{l}{Bs-AGCN$^{*}$ \cite{shi2019two}} &GCN   & 87.0 & 94.1  & - & -  \\
		\multicolumn{1}{l}{2s-AGCN$^{*}$ \cite{shi2019two}} &GCN   & 88.8 & 95.3  & - & -  \\
		\multicolumn{1}{l}{GCN-NAS(Joint\&Bone) \cite{peng2019learning}} & GCN & 89.4  & 95.7 & 37.1 & 60.1  \\
     	\multicolumn{1}{l}{DGNN \cite{shi2019skeleton}} & GCN & 89.9  & 96.1 & 36.9 & 59.6 \\
		\multicolumn{1}{l}{JB-AAGCN \cite{shi2019multi}} &GCN & 89.4  & 96.0  & 37.4 & 60.4  \\
		\multicolumn{1}{l}{MS-AAGCN \cite{shi2019multi}} &GCN & \textbf{90.0}  & \textbf{96.2 } & \textbf{37.8} & \textbf{61.0}  \\
	    \midrule
		\multicolumn{1}{l}{PeGCN$_{\text{total}}$} & GCN & 85.6 &  93.4 &  34.8 &	57.2 \\
		\midrule
		\bottomrule
	\end{tabular}
    }
	\caption{Recognition accuracies on NTU-RGB+D dataset and Kinetics-skeleton dataset. Note that, '-' indicates that the result were not reported and $^{*}$ indicates that a method is evaluated ourselves. The boldface figures denote the highest performance for each experiment. The more comprehensive comparison between PeGCN and other state-of-the-art methods are described in Appendix \ref{apx:5}}
	\label{tab:normal}
	\vspace{-4ex}
\end{wraptable}

Compared with the state-of-the art performance, PeGCN$_{\text{total}}$ produces better or comparable performance than the several methods. Js-AGCN \cite{shi2019two}, which is used as the backbone network for PeGCN$_{\text{total}}$ achieves 85.4 and 93.1 accuracies for the CS and CV protocol on NTU-RGB+D dataset. These figure are slightly lower than ours. PeGCN$_{\text{total}}$ achives 85.6 and 93.4 accuracies on the two protocal.

Nevertheless, the performances of PeGCN$_{\text{total}}$ is relatively lower than few methods such as MS-AAGCN\cite{shi2019multi}, DGNN \cite{shi2019skeleton}, GCN-NAS \cite{peng2019learning}, and AS-GCN \cite{li2019actional}.  The gap of performances between state-of-the-art methods and PeGCN can be interpreted as follows: MS-AAGCN\cite{shi2019multi} has additional attention modules (\eg Spatial , temporal, channel-wise attention) and exploiting four different modalities including joint and bone information and motion information of them. In training, batch size is twice than ours and adaptive graphs are fixed in the first 5 epochs for better learning explained in DGNN \cite{shi2019skeleton}. MS-AAGCN  achieved more top-1 accuracy than us by 5.5\%, 2.8\% and 4.0 \%  on CS, CV and Kinetics respectively. Although DGNN \cite{shi2019skeleton} has same batch size 32, it has longer training epoch as 120 while our training epoch for NTU-RGB+D is 50 and kinetics-skeleton is 65. Besides, DGNN utilizes both joint and bone information with directed acyle-graph. This leads improvement of top-1 accuracy 5.4\%, 2.7\% and 3.1\% on CS, CV and Kinetics, respectively. In other methods (such as GCN-NAS \cite{peng2019learning} and AS-GCN \cite{li2019actional}) has longer training epochs than ours and learning rate decay more frequently. 

\textbf{Experiment with noisy skeletons.} The experimental results on the skeleton-based action recognition with noisy samples clearly demonstrate the efficiency of PeGCNs in recognizing actions on noisy skeleton samples. In contrast to the other approaches that performances are rapidly degraded when the noise-levels are deepened, PeGCN shows the noise-robust action recognition performance. As shown in Table \ref{tab:ntu_cs}, PeGCN$_{\text{total}}$ achieves 84.21 and 82.20 of accuracies in the experiments with the noise level 1 and the noise level 5, respectively. The performance gap between these two figures is less than 3\%, and it is significantly lower than the other methods. Shi \etal \cite{shi2019two}, which achieves the state-of-the-art performance on the experiments with normal skeleton samples (see Table \ref{tab:normal}), produces 84.31 and 51.27 of accuracies on the noise-1 experiments, and the gap between these two accruacies is larger than 30. Js-AGCN \cite{shi2019two} achieved high accuracy that 35.1 and 57.1 for top-1 and top-5 accuracy, respectively. However, performance is dropped when noised samples are given. It recorded 23.06 on the noise-1 and 3.81 on the noise-5 experiments, and the gap between them is larger than 19. In the experiments with the noise level 10 on CS protocol in NTU-RGB+D, while the performances of other methods are all lower than 25\%, PeGCN$_{\text{total}}$ obtains 77.92\% of accuracy.

\begin{table*}[t]
	\centering
	\resizebox{\textwidth}{!}{
	\begin{tabular}{l c c c c c c c c c c}
		\toprule
		\midrule
		\multirow{2}[4]{*}{Methods} & \multicolumn{10}{c}{Noise-level} \\  \cmidrule(rl){2-11}
		& \multicolumn{2}{c}{None}  & \multicolumn{2}{c}{1} & \multicolumn{2}{c}{3}  & \multicolumn{2}{c}{5}  & \multicolumn{2}{c}{10} \\ \cmidrule(rl){2-11}
		& Top1  & Top5 & Top1  & Top5  & Top1  & Top5 & Top1  & Top5  & Top1  & Top5  \\
		\midrule
        \multicolumn{1}{l}{ST-GCN$^{*}$ \cite{yan2018spatial}} & 81.57&96.85&	73.78($\pm$0.24)&	93.74($\pm$0.13)&	57.76($\pm$0.22)&	84.08($\pm$0.24)&	42.73($\pm$0.23)&	71.52($\pm$0.25)&	17.26($\pm$0.22)&	42.54($\pm$0.32)	\\
		\multicolumn{1}{l}{Js-AGCN$^{*}$ \cite{shi2019two}}  & 86.43 & 97.28  & 76.05($\pm$0.33)&	92.09($\pm$0.16)&	54.92($\pm$0.33)&	77.56($\pm$0.23)&	35.90($\pm$0.44)&	60.67($\pm$0.27)&	8.03($\pm$0.26)&	24.91($\pm$0.21)\\

		\multicolumn{1}{l}{Bs-AGCN$^{*}$ \cite{shi2019two}}  & 87.04 &  97.48&
		79.08($\pm$0.23)&	94.03($\pm$0.10)&	60.79($\pm$0.27)&	83.75($\pm$0.26)&	44.30($\pm$0.27)&	71.80($\pm$0.20)&	18.24($\pm$0.25)&	44.07($\pm$0.16)\\
		
		\multicolumn{1}{l}{2s-AGCN$^{*}$ \cite{shi2019two}}  & \textbf{88.83} & 98.05&
        \textbf{84.31}($\pm$0.15)&	\textbf{96.73}($\pm$0.07)&	69.40($\pm$0.25)&	89.97($\pm$0.22)&	51.27($\pm$0.28)&	78.09($\pm$0.21)&	16.28($\pm$0.15)&	40.86($\pm$0.38)\\

        \multicolumn{1}{l}{Js-AAGCN$^{*}$ \cite{shi2019multi}} &87.49 & 97.45 & 80.31($\pm$0.18)&	93.62($\pm$0.12)&	65.87($\pm$0.24)&	84.58($\pm$0.23)&	51.79($\pm$0.29)&	74.06($\pm$0.24)&	21.26($\pm$0.23)&	43.89($\pm$0.41) \\

	    \multicolumn{1}{l}{3s RA-GCN$^{*}$ \cite{song2019richly}} & 85.87 & 98.10 & 72.02($\pm$0.26)&	89.89($\pm$0.20)&	45.12($\pm$0.29)&	68.79($\pm$0.33)&	25.59($\pm$0.25)&	48.71($\pm$0.42)&	6.11($\pm$0.24)&	20.55($\pm$0.31)	 \\	
		
		\multicolumn{1}{l}{2s RA-GCN$^{*}$ \cite{song2019richly}} & 85.83 & 98.19 &71.97($\pm$0.18)&	91.00($\pm$0.20)&	44.41($\pm$0.23)&	70.81($\pm$0.34)&	25.35($\pm$0.33)&	50.54($\pm$0.23)&	6.41($\pm$0.23)&	21.10($\pm$0.19)  \\
		
        \multicolumn{1}{l}{PB-GCN$^{*}$ \cite{thakkar2018part}} &86.98	&\textbf{98.25}& 77.39($\pm$0.32)&	94.67($\pm$0.15)&	56.35($\pm$0.28)&	83.03($\pm$0.12)&	37.31($\pm$0.37)&	67.87($\pm$0.36)&	11.01($\pm$0.15)&	34.13($\pm$0.24)\\
		
	    \hline
		\multicolumn{1}{l}{PeGCN$_{\text{total}}$} &  84.49 &	96.79 &
		84.21\textbf{($\pm$0.11)}&	96.72\textbf{($\pm$0.02)}&	\textbf{83.28($\pm$0.13)}&	\textbf{96.59($\pm$0.10)}&	\textbf{82.20($\pm$0.15)}&	\textbf{96.28($\pm$0.05)}&	\textbf{77.92($\pm$0.14)}&	\textbf{94.92($\pm$0.09)} \\
		\midrule
		\bottomrule
	\end{tabular}
	}
	\caption{Recognition accuracies depending on the noise level using NTU-RGB+D dataset and the CS protocol. $^{*}$ indicates that the method were trained and tested by ourselves. The boldface figures denote the highest performance for each experiment.}
	\label{tab:ntu_cs}
	\vspace{-4ex}
\end{table*}

\begin{table*}[t]
	\centering
	\resizebox{\textwidth}{!}{
	\begin{tabular}{l c c c c c c c c c c}
		\toprule
		\midrule
		\multirow{2}[4]{*}{Methods} & \multicolumn{10}{c}{Noise-level} \\  \cmidrule(rl){2-11}
		& \multicolumn{2}{c}{None}  & \multicolumn{2}{c}{1} & \multicolumn{2}{c}{3}  & \multicolumn{2}{c}{5}  & \multicolumn{2}{c}{10} \\ \cmidrule(rl){2-11}
		& Top1  & Top5 & Top1  & Top5  & Top1  & Top5 & Top1  & Top5  & Top1  & Top5  \\
		\midrule
        \multicolumn{1}{l}{ST-GCN$^{*}$ \cite{yan2018spatial}} & 88.76 & 98.83 & 83.14($\pm$0.20)&	97.44($\pm$0.08)&	69.18($\pm$0.22)&	91.95($\pm$0.15)&	54.07($\pm$0.26)&	83.30($\pm$0.28)&	23.63($\pm$0.18)&	54.58($\pm$0.34)	\\
        
        \multicolumn{1}{l}{Bs-AGCN$^{*}$ \cite{shi2019two}} & 94.12 & 99.23 &
        56.38($\pm$0.15)&	77.82($\pm$0.31)&	7.84($\pm$0.21)&	22.91($\pm$0.36)&	2.44($\pm$0.09)&	10.92($\pm$0.22)&	2.14($\pm$0.39)&	9.65($\pm$1.57) \\
        
		\multicolumn{1}{l}{Js-AGCN$^{*}$ \cite{shi2019two}}  & 94.05 & 99.08 & 85.98($\pm$0.20)&	96.34($\pm$0.13)&	68.49($\pm$0.18)&	88.03($\pm$0.20)&	51.36($\pm$0.24)&	76.61($\pm$0.24)&	17.89($\pm$0.20)&	42.20($\pm$0.29)	 \\
		
		\multicolumn{1}{l}{2s-AGCN$^{*}$ \cite{shi2019two}}  & \textbf{95.25} & 99.36 &  84.12($\pm$0.22)&	96.16($\pm$0.12)&	53.05($\pm$0.36)&	78.48($\pm$0.26)&	29.39($\pm$0.30)&	56.47($\pm$0.33)&	6.32($\pm$0.90)&	21.71($\pm$2.04) \\
		
		\multicolumn{1}{l}{Js-AAGCN$^{*}$ \cite{shi2019multi}} & 94.61 & 99.17 & 87.87($\pm$0.14)&	96.17($\pm$0.08)&	71.81($\pm$0.21)&	86.81($\pm$0.13)&	54.37($\pm$0.27)&	74.34($\pm$0.12)&	18.99($\pm$0.32)&	38.84($\pm$0.28) \\
		
		\multicolumn{1}{l}{3s RA-GCN$^{*}$ \cite{song2019richly}} & 93.51&99.30& 79.77($\pm$0.18)&	92.74($\pm$0.18)&	53.59($\pm$0.32)&	76.41($\pm$0.29)&	32.71($\pm$0.19)&	59.08($\pm$0.37)&	8.88($\pm$0.24)&	29.53($\pm$0.24) \\
		
		\multicolumn{1}{l}{2s RA-GCN$^{*}$ \cite{song2019richly}} & 92.97 & 99.28 & 79.58($\pm$0.16)&	92.72($\pm$0.11)&	53.34($\pm$0.36)&	75.09($\pm$0.24)&	32.46($\pm$0.24)&	55.84($\pm$0.32)&	8.59($\pm$0.11)&	24.98($\pm$0.20) \\

        \multicolumn{1}{l}{PB-GCN$^{*}$ \cite{thakkar2018part}} & 93.37 &	\textbf{99.37} & 80.11($\pm$0.16)&	95.16($\pm$0.12)&	54.21($\pm$0.24)&	81.5($\pm$0.19)&33.73($\pm$0.2)&	64.55($\pm$0.21)&	9.43($\pm$0.13)&	31.77($\pm$0.25) \\

	    \hline
		\multicolumn{1}{l}{PeGCN$_{\text{total}}$} &
		93.41 &	99.02 & \textbf{93.21($\pm$0.04)}&	\textbf{98.97($\pm$0.02)}&	\textbf{92.78($\pm$0.09)}&	\textbf{98.91($\pm$0.04)}&	\textbf{92.24($\pm$0.08)}&	\textbf{98.81($\pm$0.03)}&	\textbf{89.39($\pm$0.11)}&	\textbf{98.29($\pm$0.06) }\\
		\midrule
		\bottomrule
	\end{tabular}
	}
	\caption{Recognition accuracies depending on the noise level using the CV protocol of NTU-RGB+D dataset. $^{*}$ indicates that the method were trained and tested by ourselves. The boldface figures denote the highest performance for each experiment.}
	\label{tab:ntu_cv}
	\vspace{-6ex}
\end{table*}

The experimental results on the CV protocol using NTU-RGB+D dataset likewise suggest that PeGCN can provide more noise-robust performance for skeleton-based action recognition and surpass existing state-of-the-art methods. PeGCN$_{\text{total}}$ achieves state-of-the-art performance. As shown in Table \ref{tab:ntu_cv}, while PeGCN$_{\text{total}}$ achieves 99.21 and 89.39 of accuracies on the experiments with noise level 1 and 10, respectively, there is no other method that can provide over the 90\% of accuracies even in the noise level 1. Js-AAGCN$^{*}$ \cite{shi2019multi} produces 87.87 of accuracy for the noise level 1. However, the recognition performance of Js-AAGCN$^{*}$ is steeply degraded when the noise level is increased. In the experiment with noise level 10, the performance of Js-AAGCN$^{*}$ is 18.99, and it is lower than 23.63 of ST-GCN$^{*}$ \cite{yan2018spatial} which obtains 83.14 of accuracy in the experiments on the noise level 1.

\begin{table*}[t]
	\centering
	\resizebox{\textwidth}{!}{
	\begin{tabular}{l c c c c c c c c c c}
		\toprule
		\midrule
		\multirow{2}[4]{*}{Methods} & \multicolumn{10}{c}{Noise-level} \\  \cmidrule(rl){2-11}
		& \multicolumn{2}{c}{None}  & \multicolumn{2}{c}{1} & \multicolumn{2}{c}{3}  & \multicolumn{2}{c}{5}  & \multicolumn{2}{c}{10} \\ \cmidrule(rl){2-11}
		& Top1  & Top5 & Top1  & Top5  & Top1  & Top5 & Top1  & Top5  & Top1  & Top5  \\
		\midrule
		
        \multicolumn{1}{l}{ST-GCN$^{*}$ \cite{yan2018spatial}} & 31.60 & 53.68 & 22.42($\pm$0.19)&	42.91($\pm$0.23)&	8.97($\pm$0.13)&	22.24($\pm$0.20)&	3.69\textbf{($\pm$0.14)}&	11.16\textbf{($\pm$0.11)}&	0.90\textbf{($\pm$0.04)}&	3.84($\pm$0.11)	\\
		\multicolumn{1}{l}{Js-AGCN$^{*}$ \cite{shi2019two}}  & 34.39& 57.04 & 23.06($\pm$0.19)&	43.41($\pm$0.37)&	9.13($\pm$0.20)&	21.80($\pm$0.17)&	3.81\textbf{($\pm$0.14)}&	11.22($\pm$0.17)&	0.92($\pm$0.05)&	3.92($\pm$0.12)	 \\
		\multicolumn{1}{l}{Js-AAGCN$^{*}$ \cite{shi2019multi}} & \textbf{35.66} & \textbf{58.27} & 27.13($\pm$0.14)&	48.55($\pm$0.19)&	11.77($\pm$0.20)&	26.61($\pm$0.18)&	4.81($\pm$0.19)&	13.38($\pm$0.18)&	1.06($\pm$0.06)&	4.06\textbf{($\pm$0.10)}\\
		\midrule
		\multicolumn{1}{l}{PeGCN$_{\text{total}}$} & 33.78  &	56.24 & \textbf{33.34($\pm$0.13) } &	\textbf{55.84($\pm$0.09)}&	\textbf{32.45($\pm$0.12)}&	\textbf{54.78($\pm$0.09)}&	\textbf{30.90}($\pm$0.28)&	\textbf{53.37}($\pm$0.20)&	\textbf{24.04}($\pm$0.22)&	\textbf{45.41}($\pm$0.27)\\
		\midrule
		\bottomrule
	\end{tabular}
	}
	\caption{Performance comparison depending on the noise level using Kinetics-skeleton dataset.  The boldface figures denote the best performances among the listed methods. $^{*}$ indicates that the method were trained and tested by ourselves. The boldface figures denote the highest performance for each experiment.}
	\label{tab:kinetics}
	\vspace{-6ex}
\end{table*}

\subsection{Analysis and discussion}
The overall results indicate that PeGCN can provide outstanding skeleton-based action recognition robust to noisy samples compared to existing state-of-the-art methods. The accuracies of PeGCN for all noise-level on NTU-RGB+D dataset and Kinetics dataset are higher than the comparison targets. The performance gap between PeGCN and other methods is proportional to the noise level. In the experiment on noise level 10, the performances of almost methods except PeGCN are degraded over 90\% compared with the results on normal samples. In addition to the accuracies, the standard deviations also suggest that the advantage of PeGCN for noise-robust skeleton-based action recognition. In experiments for the CV protocol on NTU-RGB+D dataset, between noise levels 1 to 5, while the other methods produce the standard deviations over 0.2 usually, the range of standard deviation of the proposed method is from 0.02 to 0.11.

Interestingly, among the experimental results, RA-GCN \cite{song2019richly}, which have proposed for recognizing actions using incomplete skeletons, achieves relatively poor accuracies (Table \ref{tab:ntu_cs} and Table \ref{tab:ntu_cv}) than the other methods \cite{shi2019two,yan2018spatial,shi2019skeleton} that do not consider the skeletons with noise information. It may be caused by the difference in the definition of 'noise' on skeleton features. 
As shown in Fig. \ref{fig:noise_typ}, Song \etal \cite{song2019richly} assigned 0 to the noised joints that defined by the 'missed joints' by spatially or temporally occlusions.  However, in our experiments, the arbitrary value for the joint noising is defined randomly within the bounding box (see Fig. \ref{fig:gen_noise}). Consequently, the entire experimental results demonstrate the efficiency of PeGCN on skeleton-based action recognition with noise skeleton samples. 

\section{Conclusions}
\label{sec:5}
In this work, we have presented the noise-robust skeleton-based action recognition method based on graph convolutional networks with predictive encoding for latent space, called Predictively encoded Graph Convolutional Networks (PeGCNs). In the training step, PeGCNs learns to improve the representation ability for noise-robust skeleton-based action recognition by predicting complete samples from noisy samples on latent space. PeGCN increases the flexibility of GCNs and is more suitable for action recognition tasks using skeleton features. PeGCN is evaluated on two large-scale action recognition datasets, NTU-RGB+D and Kinetics, and it achieved the state-of-the-art performance on both of them.

\small
\bibliographystyle{splncs04}
\bibliography{egbib}

\clearpage
\setcounter{page}{1}
\setcounter{section}{1}
\setcounter{table}{0}
\setcounter{figure}{0}
 \renewcommand{\thesection}{A}
\section*{\centering{}Appendix}

\subsection{Dimensional details for the kernels on the GCN module and the autoregressive module}
\label{apx:1}

\begin{figure*}
	\vspace{-1cm}
	\begin{center}
		\centerline{\includegraphics[width=0.8\linewidth]{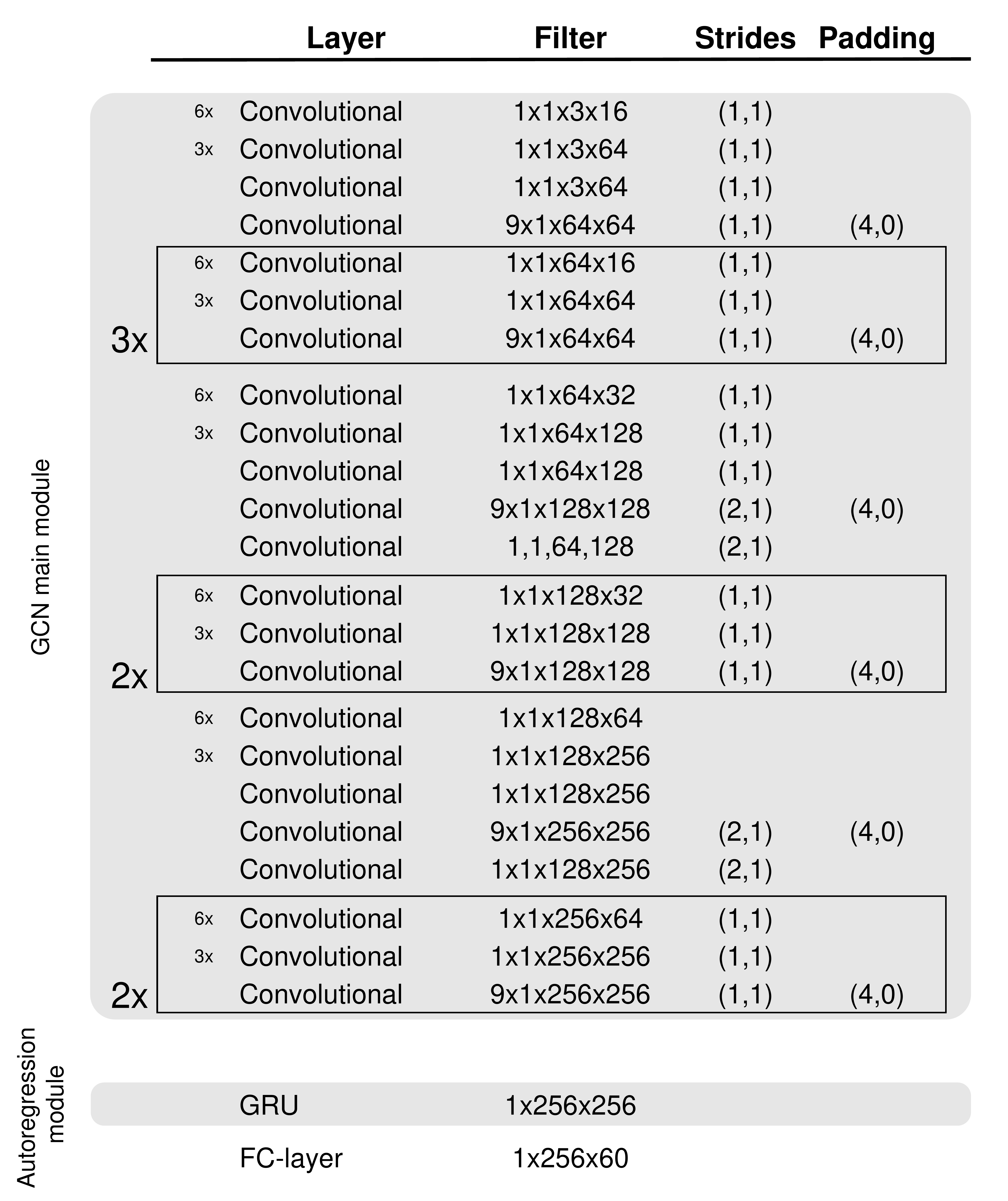}}
		\caption{Dimentionality of each layer in PeGCN model including both GCN and auto-regression modules. 
		The $'N'$x (\eg 3x or 2x) means that corresponding layer or block in solid-line  is repeating $N$ times.
		In filter column, first two figures are filter size and the last two figures are input and output dimension respectively.
		Note that, the output of FC-layer is 60 for NTU-RGB+D datset.}
	\end{center}
	\vspace{-8ex}
\end{figure*}

\newpage
\subsection{NTU-RGB+D dataset and Kinetics dataset}
\label{apx:2}
\textbf{NTU-RGB+D dataset} \cite{shahroudy2016ntu} is one of the largest dataset in skeleton based action recognition which contains around 56,000 samples in four different types including depth map, RGB video, IR image and skeleton sequence.  The samples are captured by Microsoft Kinect v2 in three different angels (-45, 0, 45) with 40 volunteers. In skeleton sequence, 3d spatial coordinates (X,Y,Z) of 25 joints are provided for each human action. The human actions are captured by one or two performers and consists of 60 indoor activities such as hand-clapping or drinking-water. \cite{shahroudy2016ntu} also provides two benchmark protocols: 1) Cross-view and 2) Cross-subject.  In cross-view protocol, samples are split into training and test set according to camera angle. Each subset contains 37,920 and 18,960 samples respectively. In cross-subject protocol, samples are split into training and test set according to subjects. Some subjects are assigned as training samples and remaining subjects are assigned as test samples. Each training and test sebsets contains 40,320 samples and 16,560 samples respectively. We follow these protocols and report the top-1 accuracy on both benchmarks.

\textbf{Kinetics-skeleton dataset} is one of the large-scale skeleton action dataset generated from Kinetics \cite{kay2017kinetics} which contains 34,000 video clips collected from Youtube to have wide variety (such as illumination change, background color) and each video clips are labeled with 400 action classes. Before estimating skeleton model from video, resolution and frame rate of video clips are converted. Skeleton model is estimated with publicly available OpenPose toolbox \cite{cao2017realtime} which gives 2d locations and 1d confidence of 18 joints. The top two person, whom has the highest average of joint confidences, in video clips are selected if multiple people are in the scene. The length of each skeleton sequence is fixed to 300 by repeating or sampling the sequence.  \cite{yan2018spatial} released this dataset (Kinetics-skeleton) which contains 240,000k samples for training set and 20,000k samples for validation set. We follow same evaluation protocol mentioned in \cite{yan2018spatial} that Top-1 and Top-5 recognition accuracies are evaluated.
\newpage

\subsection{Examples of noise skeleton samples}
\label{apx:3}
\begin{figure}
\vspace{-4 ex}
    \centering
\small
\begin{minipage}{\linewidth}
    \begin{subfigure}{\textwidth}
        \includegraphics[width=\textwidth,height=5.2cm]{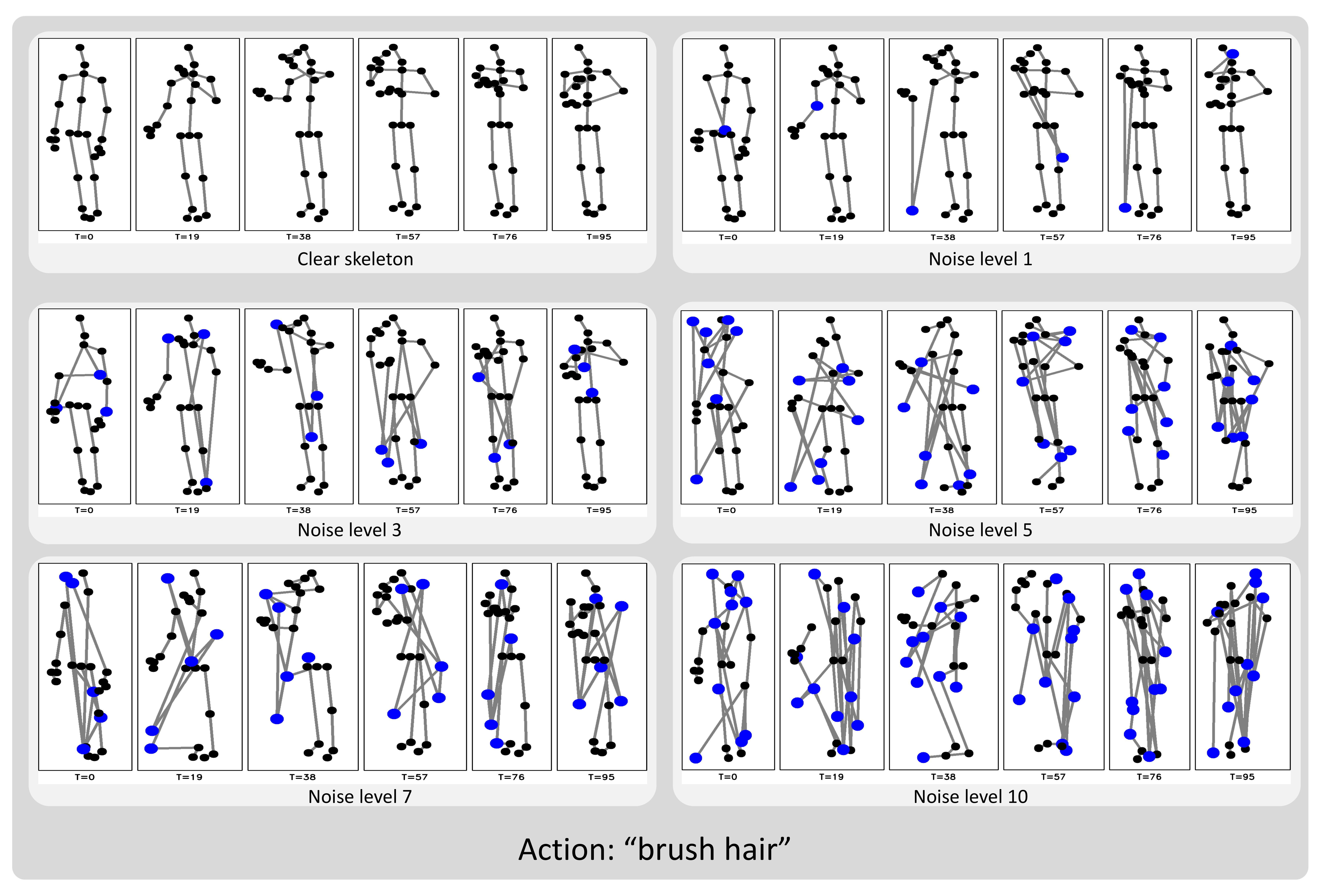}
    \end{subfigure}
    \begin{subfigure}{\textwidth}
        \includegraphics[width=\textwidth,height=5.2cm]{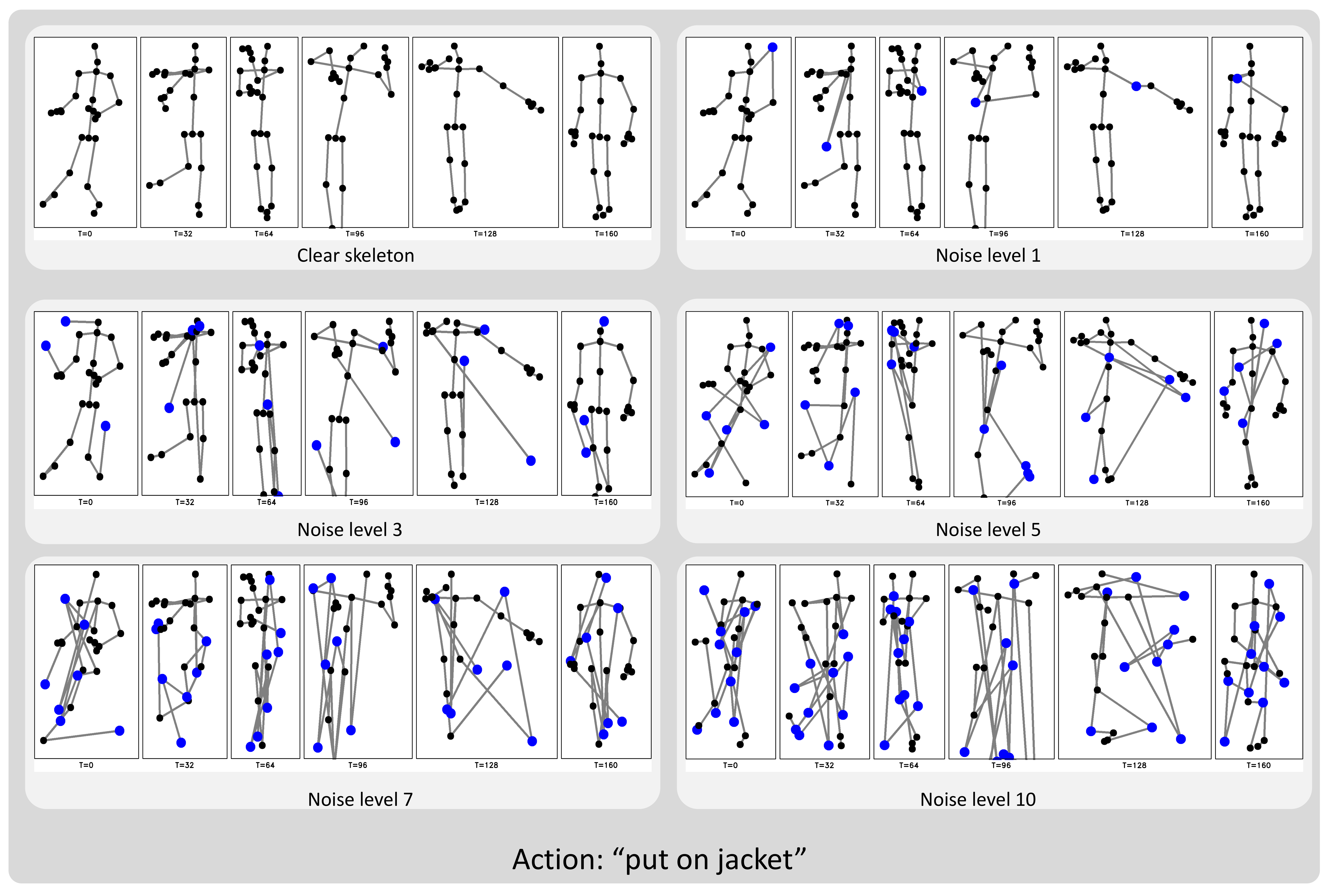}
    \end{subfigure}
    \begin{subfigure}{\textwidth}
        \includegraphics[width=\textwidth,height=5.2cm]{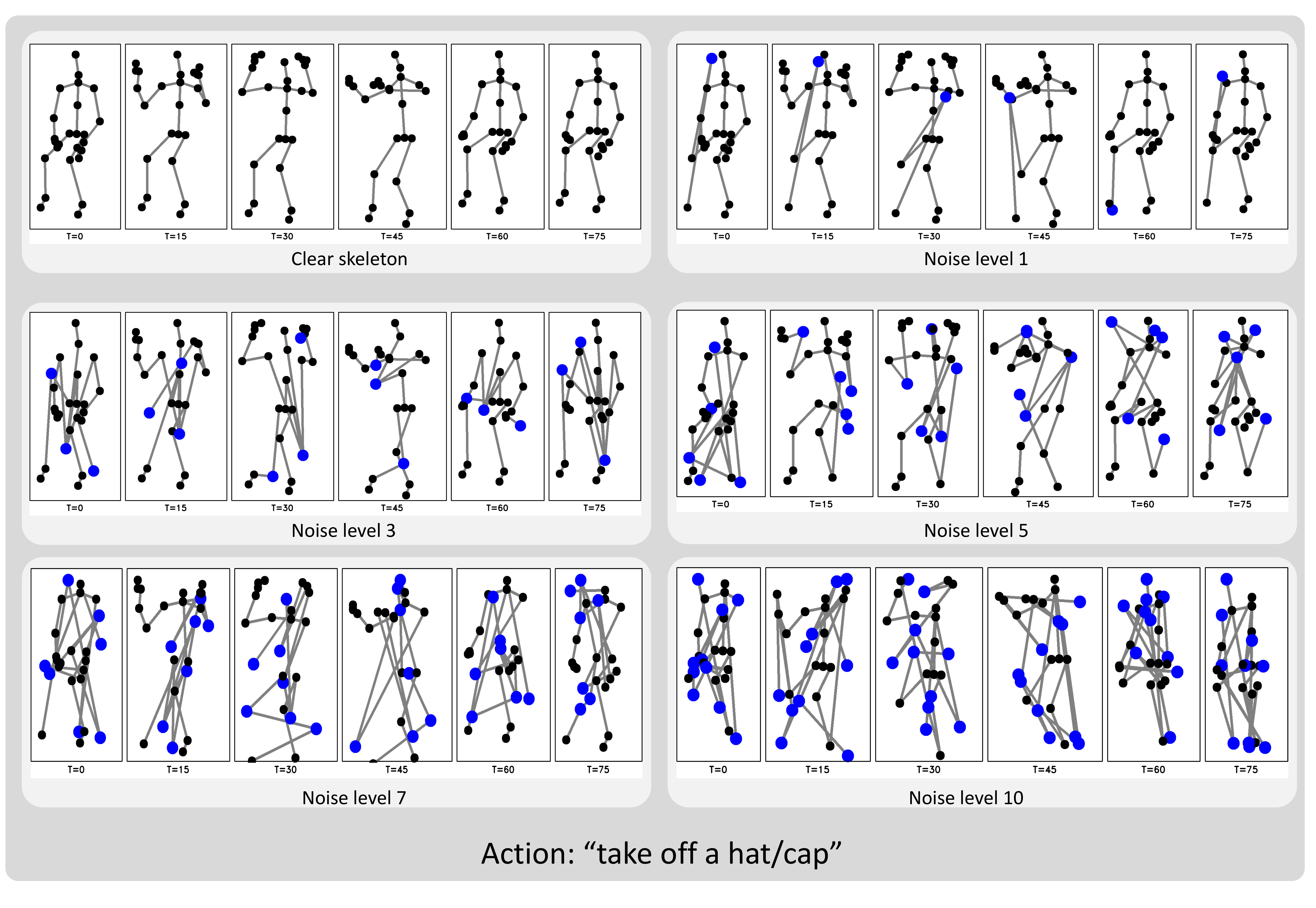}
    \end{subfigure}
    \end{minipage}
\vspace{-6ex}
\end{figure}

\newpage

\subsection{Experiment: Source codes and pre-trained models}
\label{apx:4}

All models used in our experiments are publicly available on github including our PeGCN model.
Github links are listed in Table \ref{tbl:github}. We also provides all weight files of the models via Google drive \url{https://drive.google.com/open?id=1Q-S-JAJwURPH7cy9-h25Mo0p15w_ALsb}.
Each model has multiple weight files depending on the three factors: 1) Dataset, 2) Evaluation protocol and, 3) Data-type (\eg Joint and Bone).
Details of each weight file are described in Table \ref{tbl:weights}.

\begin{table}[h]
    \caption{Publicly available source codes.}
\centering
\resizebox{0.7\textwidth}{!}{
\begin{tabular}{|l|l|} \hline
    Method & Github \\ \hline\hline

2s-AGCN \cite{shi2019two}      & \url{https://github.com/lshiwjx/2s-AGCN}		\\ \hline
MS-AAGCN \cite{shi2019multi}	&\url{https://github.com/lshiwjx/2s-AGCN} \\		 \hline
ST-GCN \cite{yan2018spatial}	&	\url{https://github.com/open-mmlab/mmskeleton}\\	\hline
PB-GCN \cite{thakkar2018part}	& \url{https://github.com/kalpitthakkar/pb-gcn} \\	 \hline
2s-RA-GCN \cite{song2019richly}&	\url{https://github.com/yfsong0709/RA-GCNv1}\\ \hline
3s-RA-GCN \cite{song2019richly}	& \url{https://github.com/yfsong0709/RA-GCNv1} \\ \hline \hline 
PeGCN	& \url{https://github.com/andreYoo/PeGCNs}	\\ \hline
\end{tabular}
}
\vspace{0.5ex}
\label{tbl:github}
\end{table}

\begin{table}[h]
\caption{ All weight files for each GCN method. Note that, - symbol in Protocol column indicates that nothing is determined and $^{*}$ symbol indicates that models are trained by ourselves and the others are downloaded from official github page.}
\centering
\resizebox{0.7\textwidth}{!}{

\begin{tabular}{|l|l|l|l|l|}
\hline
\multicolumn{1}{|c|}{Method}& 
\multicolumn{1}{c|}{Dataset}& 
\multicolumn{1}{c|}{Protocol}& 
\multicolumn{1}{c|}{Type}& 
\multicolumn{1}{c|}{Weight file} \\ \hline
\multirow{5}{*}{2s-AGCN \cite{shi2019two}} 
                         & \multirow{4}{*}{NTU-RGB+D}         & \multirow{2}{*}{Cross-View}    & Joint &ntu\_cv\_agcn\_joint\\ \cline{4-5} 
                         &                                    &                                & Bone  &ntu\_cv\_agcn\_bone\\ \cline{3-5} 
                         &                                    & \multirow{2}{*}{Cross-Subject} & Joint &ntu\_cs\_agcn\_joint\\ \cline{4-5} 
                         &                                    &                                & Bone  &ntu\_cs\_agcn\_bone\\ \cline{2-5} 
                         & \multirow{2}{*}{Kinetics-Skeleton} & \multicolumn{1}{c|}{-}         & Joint &ki\_agcn\_joint\\ \cline{3-5}
                         &                                    & \multicolumn{1}{c|}{-}         & Bone  &-            \\ \hline

\multirow{6}{*}{MS-AAGCN \cite{shi2019multi}} 
                         & \multirow{4}{*}{NTU-RGB+D}         & \multirow{2}{*}{Cross-View}    & Joint &ntu\_cv\_aagcn\_joint\\ \cline{4-5} 
                         &                                    &                                & Bone  &-             \\ \cline{3-5} 
                         &                                    & \multirow{2}{*}{Cross-Subject} & Joint &ntu\_cs\_aagcn\_joint\\ \cline{4-5} 
                         &                                    &                                & Bone  &-             \\ \cline{2-5} 
                         & \multirow{2}{*}{Kinetics-Skeleton} & \multicolumn{1}{c|}{-}         & Joint &ki\_aagcn\_joint\\ \cline{3-5} 
                         &                                    & \multicolumn{1}{c|}{-}         & Bone  &-            \\ \hline
                         
\multirow{3}{*}{ST-GCN \cite{yan2018spatial}} 
                         & \multirow{2}{*}{NTU-RGB+D}         & \multirow{1}{*}{Cross-View}    & Joint &st\_gcn.ntu-xview\\ \cline{4-5} 
                         \cline{3-5} 
                         &                                    & \multirow{1}{*}{Cross-Subject} & Joint &st\_gcn.ntu-xsub\\ \cline{4-5} 
                         \cline{2-5} 
                         & \multirow{1}{*}{Kinetics-Skeleton} & \multicolumn{1}{c|}{-}         & Joint &st\_gcn.kinetics\\ \cline{3-5} 
                          \hline

\multirow{2}{*}{PB-GCN \cite{thakkar2018part}} 
                         & \multirow{2}{*}{NTU-RGB+D}         & \multirow{1}{*}{Cross-View}    & Joint &crossview\_weights\\ \cline{4-5} 
                         \cline{3-5} 
                         &                                    & \multirow{1}{*}{Cross-Subject} & Joint &crosssubject\_weights\\ \cline{4-5} 
                         \cline{2-5} 
                         \cline{3-5} 
                         \hline
\multirow{2}{*}{2s-RA-GCN \cite{song2019richly}} 
                         & \multirow{2}{*}{NTU-RGB+D}         & \multirow{1}{*}{Cross-View}    & Joint &3304\_2s\_RA-GCN\_NTUcv.pth\\ \cline{4-5} 
                         \cline{3-5} 
                         &                                    & \multirow{1}{*}{Cross-Subject} & Joint &3302\_2s\_RA-GCN\_NTUcs.pth\\ \cline{4-5} 
                         \cline{2-5} 
                         \cline{3-5} 
                         \hline
                         
\multirow{2}{*}{3s-RA-GCN \cite{song2019richly}} 
                         & \multirow{2}{*}{NTU-RGB+D}         & \multirow{1}{*}{Cross-View}    & Joint &3303\_3s\_RA-GCN\_NTUcv.pth\\ \cline{4-5} 
                         \cline{3-5} 
                         &                                    & \multirow{1}{*}{Cross-Subject} & Joint &3301\_3s\_RA-GCN\_NTUcs.pth\\ \cline{4-5} 
                         \cline{2-5} 
                         \cline{3-5} 
                         \hline

\multirow{6}{*}{PeGCN} 
                         & \multirow{4}{*}{NTU-RGB+D}         & \multirow{2}{*}{Cross-View}                &  \multirow{2}{*}{Joint} &ntu\_cv\_magcn\_joint\_gcn\\ \cline{5-5} 
                         &                                    &                                            &       &ntu\_cv\_magcn\_joint\_ar\\ \cline{3-5} 
                         &                                    & \multirow{2}{*}{Cross-Subject}             & \multirow{2}{*}{Joint}  &ntu\_cs\_magcn\_joint\_gcn\\ \cline{5-5} 
                         &                                    &                                            &       &ntu\_cs\_magcn\_joint\_ar\\ \cline{2-5} 
                         & \multirow{2}{*}{Kinetics-Skeleton} & \multicolumn{1}{c|}{\multirow{2}{*}{-}}    & \multirow{2}{*}{Joint}  &ki\_magcn\_joint\_gcn\\  \cline{5-5}
                         &                                    &  \multicolumn{1}{c|}{}                     &       &ki\_magcn\_joint\_gr\\ \hline
        
\end{tabular}
}\vspace{0.5ex}
\label{tbl:weights}
\end{table}

\newpage
\subsection{Extended comparison on skeleton-based action recognition performance using normal skeletons on NTU-RGB+D dataest}
\label{apx:5}
\begin{table*}[h]
\vspace{-6ex}
	\centering
		\caption{Performance comparison on NTU-RGB+D dataset and Kinetics-skeleton dataset. '-' indicates that the result were not reported. $^{*}$ indicates that model is trained by ourselves and figures in parentheses means reported accuracy. The boldface figures denote the highest performance for each experiment.}
	\resizebox{\textwidth}{!}{
	\begin{tabular}{l c c c c c c c c}
		\toprule
		\midrule
		\multirow{2}[4]{*}{Methods}  & 	
		\multirow{2}[4]{*}{Year}  & 	

		\multirow{2}[4]{*}{Architecture}  & 
		
		\multicolumn{2}{c}{NTU-CS} & 
		
		\multicolumn{2}{c}{NTU-CV}  & 
		
		\multicolumn{2}{c}{Kinetics-skeleton}  
		
		\\ \cmidrule(rl){4-5} \cmidrule(rl){6-7} \cmidrule(l){8-9}& 
		& & Top1  & Top5  & Top1  & Top5 & Top1  & Top5   \\
		\cmidrule(l){1-9}
        \multicolumn{1}{l}{Fenture Enc \cite{fernando2015modeling}} & 2015&Hand  & - & - & - & - & 14.9 & 25.8  \\
        \hline
		\multicolumn{1}{l}{HBRNN \cite{du2015hierarchical}}  &2015  & RNN & 59.1 & - & 64.0 &- &- &- \\
		\multicolumn{1}{l}{Deep LSTM \cite{shahroudy2016ntu}}&2016 & LSTM & 60.7 & - & 67.3 &- & 16.4 & 35.3\\
		\multicolumn{1}{l}{ST-LSTM \cite{liu2016spatio}}  &2016    & LSTM & 69.2 & - & 77.7 &- &- &- \\
		\multicolumn{1}{l}{STA-LSTM \cite{song2017end}}  &2017     & LSTM & 73.4 & - & 81.2 &- &- &- \\
		\multicolumn{1}{l}{VA-LSTM \cite{zhang2017view}} &2017     & LSTM & 80.7 & - & 88.8 &- &- &- \\
		\midrule
		\multicolumn{1}{l}{TCN \cite{kim2017interpretable}}  &2017   & CNN & 74.3 & - & 83.1 &-& 20.3 & 40.0 \\
		\multicolumn{1}{l}{Clips+CNN+MTLN \cite{ke2017new}} & 2017&  CNN & 79.6&- & 84.8  &- &  - &- \\
		\multicolumn{1}{l}{Synthesized CNN \cite{liu2017enhanced}}& 2017 & CNN &  80.0 &- & 87.2&-  &- &- \\
		\multicolumn{1}{l}{3scale ResNet152 \cite{li2017skeleton}} & 2017& CNN & 85.0 & - &  92.3  &-  &- &- \\
 	    \midrule
 		\multicolumn{1}{l}{DPRL+GCNN \cite{tang2018deep}} & 2018 &GCN &        83.6   &-   & 89.8 & - &-   &- \\
	    \multicolumn{1}{l}{AGC-LSTM(Joint\&Part) \cite{si2019attention}} & 2019 &GCN+LSTM &89.2 &- &95.0 &- &-&-  \\
		\multicolumn{1}{l}{AS-GCN \cite{li2019actional}}& 2019  &GCN &     86.8  & -   &   94.2  & -   & 34.8 & 56.5   \\
		\multicolumn{1}{l}{ST-GCN$^{*}$ \cite{yan2018spatial}}& 2018  & GCN & 81.6(81.5) & 96.9 & 88.8(88.3)& 98.8 & 31.6(30.7) &53.7(52.8) \\
		\multicolumn{1}{l}{2s RA-GCN$^{*}$ \cite{song2019richly}}& 2019 & GCN &  85.8(85.8) & 98.2 & 93.0(93.0) & 99.3  & - &- \\
		\multicolumn{1}{l}{3s RA-GCN$^{*}$ \cite{song2019richly}}& 2019 & GCN &  85.9(85.9) & 98.1 & 93.5(93.5) & 99.3  & - &- \\
        \multicolumn{1}{l}{PB-GCN$^{*}$ \cite{thakkar2018part}} & 2018 & GCN  &  87.0(87.5) & 98.3 & 93.4(93.2) & 99.4 & - & -\\
        \multicolumn{1}{l}{Js-AGCN$^{*}$ \cite{shi2019two}}& 2019 &GCN   & 85.4 & 97.3  & 93.1(93.7)   & 99.08  & 34.4(35.1) & 57.1(57.1) \\
		\multicolumn{1}{l}{Bs-AGCN$^{*}$ \cite{shi2019two}}& 2019  &GCN   & 87.0 & 97.5  & 94.1(93.2) &  99.23   & 34.1(33.3) &57.0(55.7)  \\
		\multicolumn{1}{l}{2s-AGCN$^{*}$ \cite{shi2019two}}& 2019 &GCN   & 88.8(88.5) & 98.1  & 95.3(95.1)  & 99.4 & 36.8(36.1) & 59.2(58.7)  \\
		\multicolumn{1}{l}{GCN-NAS(Joint\&Bone) \cite{peng2019learning}}& 2019 &GCN & 89.4 &- &95.7 &- &37.1& 60.1  \\
     	\multicolumn{1}{l}{DGNN \cite{shi2019skeleton}}& 2019 & GCN & 89.9  &- & 96.1 &- & 36.9 & 59.6 \\
		\multicolumn{1}{l}{JB-AAGCN \cite{shi2019multi}}& 2019 &GCN & 89.4 & - & 96.0 &- & 37.4 & 60.4  \\
		\multicolumn{1}{l}{MS-AAGCN \cite{shi2019multi}}& 2019 &GCN & \textbf{90.0} & - & \textbf{96.2} &- & \textbf{37.8} & \textbf{61.0}  \\
	    \midrule
		\multicolumn{1}{l}{PeGCN$_{\text{total}}$}& 2020 & GCN & 85.6 &	96.79&  93.41&	99.02&  34.8&	57.24 \\
		\midrule
		\bottomrule
	\end{tabular}
    	}
	\label{tab:normal:apx}
	\vspace{-4ex}
\end{table*}

\subsection{Additional comparison on skeleton-based action recognition performance using noisy skeletons on Kinetics-skeleton dataset}
\label{apx:6}
\begin{table*}
\vspace{-8ex}
	\centering
	\caption{ Performance comparison depending on the noise level using Kinetics-skeleton dataset. The boldface figures denote the highest performance for each experiment. }

	\resizebox{\textwidth}{!}{
	\begin{tabular}{l c c c c c c c c c c}
		\toprule
		\midrule
		\multirow{2}[4]{*}{Methods} & \multicolumn{10}{c}{Noise-level} \\  \cmidrule(rl){2-11}
		& \multicolumn{2}{c}{None}  & \multicolumn{2}{c}{1} & \multicolumn{2}{c}{3}  & \multicolumn{2}{c}{5}  & \multicolumn{2}{c}{10} \\ \cmidrule(rl){2-11}
		& Top1  & Top5 & Top1  & Top5  & Top1  & Top5 & Top1  & Top5  & Top1  & Top5  \\
		\midrule
		
		\multicolumn{1}{l}{Js-AGCN$^{*}$ \cite{shi2019two}} & 34.39& 57.04 & 23.06($\pm$0.19)&	43.41($\pm$0.37)&	9.13($\pm$0.20)&	21.80($\pm$0.17)&	3.81($\pm$0.14) &	11.22($\pm$0.17)&	0.92($\pm$0.05)&	3.92($\pm$0.12)	 \\
        \multicolumn{1}{l}{Bs-AGCN$^{*}$ \cite{shi2019two}} & 34.11&	56.97& 24.01($\pm$0.21)&	44.48($\pm$0.24)&	10.03($\pm$0.14)&	23.05($\pm$0.21)&	3.99($\pm$0.15)&	11.34($\pm$0.15)&	0.82($\pm$0.08)&	3.17($\pm$0.07) \\
        \multicolumn{1}{l}{2s-AGCN$^{*}$ \cite{shi2019two}} & 36.77& 	59.24 &  28.27($\pm$0.13)&	50.11($\pm$0.12)&	12.92($\pm$0.12)&	28.18($\pm$0.25)&	5.32($\pm$0.18)&	14.52($\pm$0.16)&	1.09($\pm$0.06)&	4.15($\pm$0.10) \\

		\midrule
		\multicolumn{1}{l}{PeGCN$_{\text{total}}$} & 33.78  &	56.24 & \textbf{33.34($\pm$0.13) } &	\textbf{55.84($\pm$0.09)}&	\textbf{32.45($\pm$0.12)}&	\textbf{54.78($\pm$0.09)}&	\textbf{30.90}($\pm$0.28)&	\textbf{53.37}($\pm$0.20)&	\textbf{24.04}($\pm$0.22)&	\textbf{45.41}($\pm$0.27)\\
		\midrule
		\bottomrule
	\end{tabular}
	}
	\label{tab:tmp}
	\vspace{-4ex}
\end{table*}

\end{document}